\RequirePackage{fix-cm}
\documentclass[smallextended]{svjour3}       
\smartqed  
\usepackage{graphicx}
\usepackage{caption}
\usepackage{subcaption}
\usepackage{xcolor}

\usepackage{algorithm}
\usepackage{algorithmic}

\usepackage{amssymb}
\usepackage{amsmath}
\usepackage{mathabx}

\begin{document}

\title{Which is the best model for my data? 
}


\author{Gonzalo N{\'a}poles  \and  Isel Grau \and {\c C}i{\c c}ek G{\"u}ven \and Or{\c c}un {\"O}zdemir \and Yamisleydi Salgueiro}


\institute{G. N{\'a}poles  \at
              Department of Cognitive Science \& Artificial Intelligence, Tilburg University, The Netherlands  \\
              \email{g.r.napoles@uvt.nl}           
        \and
           I. Grau \at
              Information Systems Group, Eindhoven University of Technology, The Netherlands
        \and
           {\c C}. G{\"u}ven \at
              Department of Cognitive Science \& Artificial Intelligence, Tilburg University, The Netherlands
        \and
           O. {\"O}zdemir \at
              Department of Cognitive Science \& Artificial Intelligence, Tilburg University, The Netherlands
        \and
           Y. Salgueiro \at
              Department of Computer Science, Universidad de Talca, Campus Curic\'o, Chile
}

\date{Received: date / Accepted: date}

\maketitle

\begin{abstract}
In this paper, we tackle the problem of selecting the optimal model for a given structured pattern classification dataset. In this context, a model can be understood as a classifier and a hyperparameter configuration. The proposed meta-learning approach purely relies on machine learning and involves four major steps. Firstly, we present a concise collection of 62 meta-features that address the problem of information cancellation when aggregation measure values involving positive and negative measurements. Secondly, we describe two different approaches for synthetic data generation intending to enlarge the training data. The former approach produces completely synthetic classification problems, while the latter is a generative model that produces extra training instances from a limited pool of real-world problems. Thirdly, we fit a set of pre-defined classification models for each classification problem while optimizing their hyperparameters using grid search. The goal is to create a meta-dataset such that each row denotes a multilabel instance describing a specific problem. The features of these meta-instances denote the statistical properties of the generated datasets, while the labels encode the grid search results as binary vectors such that best-performing models are positively labeled. Finally, we tackle the model selection problem with several multilabel classifiers, including a Convolutional Neural Network designed to handle tabular data. The simulation results show that our meta-learning approach can correctly predict an optimal model for 91\% of the synthetic datasets and for 87\% of the real-world datasets. Furthermore, we noticed that most meta-classifiers produced better results when using our meta-features. Overall, our proposal differs from other meta-learning approaches since it tackles the algorithm selection and hyperparameter tuning problems in a single step. Toward the end, we perform a feature importance analysis to determine which statistical features drive the model selection mechanism.
\keywords{AutoML \and Algorithm selection \and Hyperparameter tuning}

\end{abstract}

\section{Introduction}
\label{section:introduction}

The algorithm selection problem was first introduced in \cite{Rice1976} and continues to be an open challenge. This problem consists of learning a meta-model able to relate the properties of the data with the performance of algorithms. In that way, we could predict which algorithm is more likely to perform the best given a new dataset. This approach can be considered a way to alleviate the practical limitations imposed by the \textit{no-free lunch theorem}.

Algorithm selection is a pivotal piece of the Automated Machine Learning (AutoML) field \cite{Hutter2019}, which also includes cleaning the data and pre-processing steps, selecting or extracting relevant features, recommending the best algorithm fitting the problem, optimizing the hyperparameters, and post-processing the results whenever applicable. Some relevant AutoML frameworks supporting the creation of intelligent systems include AutoWEKA \cite{Thornton2013}, Auto-sklearn \cite{feurer-neurips15a}, Auto-PyTorch \cite{Zimmer2021} and AutoKeras \cite{jin2019auto}.

As a part of the AutoML pipeline, algorithm selection \cite{Hospedales2021}\cite{Khan2020} typically relies on a supervised meta-model that relates meta-features with the performance of candidate prediction algorithms. These features could be as simple as the number of features, instances or decision classes, or based on statistics or information-theoretic principles such as mutual information or the eigenvalues. When it comes to hyperparameter tuning, the predominant approach is optimization-based, with constrained Bayesian optimization being a suitable approach \cite{Hutter2019} to fulfill memory and time constraints.

Despite the progress concerning algorithm selection and hyperparameter tuning reported in the literature, some major challenges persist. Firstly, existing meta-features often hide relevant information as they aggregate negative and positive values that might cancel each other. Secondly, the meta-classifiers are usually built on a limited collection of datasets, which might affect their generalization capabilities in practice. Thirdly, finding the optimal hyperparameter setting continues to be a computationally demanding, energy-unfriendly process as it involves optimizing a loss function across different training and validation sets. Furthermore, for a given problem, there could be several optimized models reporting the highest performance, yet correctly recognizing one of them would solve the model selection problem. Finally, there is a limited understanding of the relevance of meta-features on the performance of meta-classifiers focusing on algorithm selection.

This paper presents a meta-learning approach to tackle the above-mentioned issues when selecting the optimal model for a given structured classification problem. The main contribution of our approach is that it tackles the algorithm selection and hyperparameter tuning problems in a single step by predicting the best model for a given problem. In this context, a model can be understood as a machine learning algorithm using a specific hyperparameter setting. The added value of selecting the model instead of selecting the algorithm is that a subsequent hyperparameter tuning procedure is not required. Furthermore, we approach the model selection problem as a multilabel learning problem where each problem can be associated with several optimal models simultaneously, even when the goal is effectively recognizing only one of them. The steps describing our contribution are explained below.

\begin{itemize}
    \item First, we present a concise collection of 62 meta-features that address the problem of information cancellation when aggregation measure values involving positive and negative measurements.
    
    \item Second, we describe two different approaches for synthetic data generation with the aim of enlarging the training data. The first approach relies on the \texttt{make\_classification} function from \texttt{Sklearn} to generate 10,000 random classification problems with different complexity levels (as defined by the class separation, imbalance ratio, and the number of non-informative features, among others). The second approach uses a conditional Generative Adversarial Network (GAN) for producing 1,000 synthetic instances from a pool of 52 real-world classification datasets.
     
    \item Third, we fit a pre-defined collection of models (that might belong to different families) on each dataset and determine the best-performing models on the validation set. The aim is to create a meta-dataset where each instance represents a classification problem described by a set of meta-features and associated with a multilabel output encoding the performance grid (resulting from the tuning step) as a binary vector. In this approach, best-performing models are positively labeled, while those models that did not reach the optimal performance are negatively labeled.

    \item Fourth, we tackle the model selection problem by fitting a Convolutional Neural Network (CNN) on each meta-dataset. Notice that the generation step produces two meta-datasets: one describing synthetically generated problems and the other containing meta-instances obtained with a GAN model from real-world datasets. In both cases, the overall goal is to accurately predict an optimal model for a given dataset, instead of predicting whether \textit{each} model would perform optimally or not. However, the learning problem continues to be multilabel by nature.
 
\end{itemize}

The simulation results show that our meta-learning approach accurately predicted an optimal model for 91\% of the completely synthetic datasets and 87\% of the real-world datasets. This measure is referred to as \textit{hit rate} and quantifies how often the most likely positive model predicted by the meta-classifier was indeed an optimal model according to the ground truth. Let us reiterate that the novelty of our proposal is that it tackles the algorithm selection and hyperparameter tuning problems in a single step. Moreover, it was found that the tested meta-classifiers systematically reported larger hit rates when using the proposed meta-features compared to other state-of-the-art meta-features. Toward the end, we conducted a feature importance analysis to determine which statistical features drive the model selection mechanism. Our findings suggest that the correlation between the features, the distribution of decision classes, the cluster-based measures and the performance of weaker classifiers are reliable proxies for model selection.

The remainder of this paper is as follows. Sect.~\ref{sec:literature} revises the state-of-the-art concerning meta-learning approaches. Sect.~\ref{sec:meta-features} presents a compact set of meta-features that handle information aggregation more efficiently when operating with signed values. Sect.~\ref{sec:generation} describes two approaches for generating synthetic data to be used for enlarging the training sets. Sect.~\ref{sec:proposal} introduces the proposed multilabel approach for model selection, which uses a CNN model as the core meta-classifier. Sect.~\ref{sec:simulations} conducts a set of numerical simulations involving both synthetic and real-world problems and performs a feature importance study. Sect.~\ref{sec:conclusions} elaborates on the limitations of our method, while discussing improvement points to be addressed in future studies.

\section{Related work}
\label{sec:literature}


After the seminal work of Rice \cite{Rice1976}, several approaches for mapping a set of meta-features to a space of candidate algorithms have been proposed. According to Khan et al. \cite{Khan2020}, the main characteristics that differentiate these approaches are the types of meta-features describing the problems, the type of meta-learning algorithm used for discovering the mapping, and the type of output of the meta-learner.

Regarding the meta-features, it is a common practice to include statistical measures that are computationally feasible to calculate for a given dataset (such as the number of features, the number of decision classes, the proportion of missing values, or the standard deviation). Overall, meta-features are assumed to hold an intrinsic relationship with the performance of the algorithms being considered. An overview of meta-features for classification tasks can be consulted in \cite{vanschoren2018meta} and \cite{rivolli2018characterizing,rivolli2022meta}, where the meta-features are classified into groups, namely simple (general), statistical, information-theoretic, model-based, land-marking, and others. In an attempt to homogenize the choice of meta-features \cite{pinto2016} propose a framework to systematically generate meta-features. More recently, the authors in \cite{JMLR:meta-features} propose a meta-feature extractor package. The characterization and standardization of these measures help the comparison of meta-learning experiments in a reproducible manner \cite{rivolli2018characterizing}, \cite{JMLR:meta-features}. In a different direction, \cite{rakotoarison2022learning} learn extra meta-features as linear combinations of the manually designed meta-features of the literature. In \cite{rivolli2022meta}, the authors summarize the existing tools to extract meta-features.

The most common meta-learning approach is based on the similarity between a given problem and others encoded in the meta-dataset \cite{kalousis1999noemon,brazdil2003ranking,SONG20122672,Guangtao2014}. For example, the work proposed in \cite{SONG20122672} uses 84 datasets and 17 algorithms to provide a ranked list of the best-performing algorithms for similar meta-instances. Providing the results in the form of a ranked list of algorithms comes from the direct mapping of the top nearest neighbors to the algorithms space, based on a measure of performance, in this case, a combination of accuracy and run-time. The main advantage of using kNN as a meta-learner is its possibility for extension since all new meta-instances can be added without retraining the meta-learning model. However, the results of the meta-learning are sensitive to the choice of the value for the number of neighbors $k$ \cite{ZHU2018171}.

Another meta-learning approach for algorithm selection is rule-based models. Early work published in \cite{Brazdil1994} proposes a C4.5 decision tree trained on the information of 22 datasets characterized by statistical meta-features (e.g., the mean entropy of the features) and mapped to a pool of 22 algorithms. More recently, \cite{ALI2006119} develops a wider study including 100 classification problems and eight candidate algorithms, using C5.0 as a decision tree learning algorithm for the meta-learner. A clear advantage of decision trees is their interpretability, which translates into the possibility of analyzing the rules that lead to the choice of an algorithm. However, decision trees fail to outperform other approaches in terms of the accuracy of the meta-learner model \cite{Khan2020}.

Additionally, an interesting diverging idea is to model meta-learning as a link prediction problem in a graph structure \cite{ZHU2018171}. The authors test their method on 131 datasets and 21 classification algorithms, outperforming the baseline kNN approach. Similar to kNN, the output of the meta-learner is a ranking of the best-performing algorithms. Finally, other alternatives include modeling the problem as a regression problem \cite{reif2014automatic,bensusan2001estimating}, where the meta-learner outputs an estimate of the accuracy of the best model directly; or as a classification after clustering problem \cite{wang2015improved}. 

One of the primary drawbacks found in the revised literature is the limited number of datasets used for the experimentation, with the widest study considering 131 datasets \cite{ZHU2018171}. There is no clear guideline concerning the number of datasets that should be considered \cite{Khan2020}. However, we argue that synthetic data generation can help increase the number of training instances, thus leading to improved results. A second aspect that draws our attention is that most approaches are either lazy or rule-based learners. Interestingly, we are unaware of neural network models used as meta-learners, possibly due to the limited number of training data used in previous studies. A third observation is that the solution is often limited to algorithm selection without considering its hyperparameters, as this issue is tackled in a separate step. Along the same line, most studies output a ranked list of algorithms, while for a given problem, several fully optimized models could report the highest performance. Only a few published works \cite{Guangtao2014,Brazdil1994,ALI2006119} include the possibility of outputting several winners, modeling the problem as a multilabel classification. Finally, only the rule-based models attempt to provide insights into the relevance and relations of meta-features with the best-performing algorithms.

\section{Statistical meta-features for classification datasets}
\label{sec:meta-features}

 
The first step in our methodology is to define a set of informative meta-features to describe tabular structured classification problems. As mentioned in the previous section, the literature includes several packages devoted to this task. A common limitation of existing meta-features is that they often compute the mean over negative and positive values that might cancel each other. In this section, we describe a concise set of features that outperform the ones in the \texttt{pymfe} package when it comes to model selection. 

The first meta-features used to describe the classification problems are the number of features (\texttt{n\_features}), number of instances (\texttt{n\_instances}), the number of decision classes (\texttt{n\_classes}), the number of normally distributed features (\texttt{n\_normal\_features}), and the maximum eigenvalue (\texttt{max\_eigenvalue}). The latter provides useful information about the amount of variance captured through a linear combination of features.

Since machine algorithms trained on imbalanced data are biased towards the dominant decision class, understanding class imbalance using entropy \cite{vanschoren2018meta} is crucial. The class entropy (\texttt{class\_entropy}) is given by:

\begin{equation}
\label{eq:class_entropy}
\begin{split}
E(y) = -\sum_i p(y(i))~\text{log}~p(y(i))
\end{split}
\end{equation}

Other meta-features describing the target feature calculate the occurrence of each decision class and divide it by the number of instances. To obtain a richer representation, we compute the minimum, maximum, mean and standard deviation values for the class distribution. These values are represented as \texttt{min\_class\_distribution}, \texttt{max\_class\_distribution}, \texttt{mean\_class\_distribution}, \texttt{std\_class\_distribution}, respectively. 



In order to quantify the association between the features, we can rely on the absolute Pearson's correlation coefficient. By doing that, we compute the minimum, maximum, mean and standard deviation values across all feature pairs. These meta-features are represented as \texttt{min\_corr\_ff}, \texttt{max\_corr\_ff}, \texttt{mean\_corr\_ff} and \texttt{std\_corr\_ff}, respectively.

We can also compute the mutual information between the problem features and the target \cite{Ross2014}. Let $x \in X$ be a continuous random variable with $X$ being the set of all possible variables, $y$ the categorical target variable, and $p(y)$ the discrete probability function attached to the target variable. If no confusion arises, we will use $\Phi$ to denote the set of variable-target pairs. Equation \eqref{eq:mutual_information} shows how to calculate the mutual information, 

\begin{equation}
\label{eq:mutual_information}
\begin{split}
M(x,y) = E(y) + E(x) - E(y,x)
\end{split}
\end{equation}

\noindent such that $E(y)$ is given Equation \eqref{eq:class_entropy}, $E(x)$ is given by $-\int p(x(i))$ $~\text{log}~p(x(i)) \,dx$ and $E(y,x)$ is given by $-\sum_i \mu(y(i),x(i))~\text{log}~\mu(y(i),x(i))$. The previous expression can be simplified as follows:

\begin{equation*}
\label{eq:mutual_information_e}
M(x,y) = E(x) + \sum_i \mu(y(i),x(i))~\text{log}~\mu(x(i)|y(i))
\end{equation*}

\noindent where $\mu(x(i))$ denotes the probability density for sampling $x(i)$ irrespective of the decision class, while $\mu(x(i)|y(i)) = \mu(y(i),x(i))) / p(y(i))$ with $\mu(y(i),x(i)))$ being the joint probability distribution for the continuous variable and the categorical one denoting the decision class. Based on this measure, we compute the minimum (\texttt{min\_corr\_fc}), maximum (\texttt{max\_corr\_fc}), average (\texttt{mean\_corr\_fc}) and standard deviation (and \texttt{std\_corr\_fc}) mutual information values across all pairs of features and decision classes.

Another interesting measure to be explored refers to the fuzzy partition coefficient (\texttt{fuzzy\_part\_coeff}) obtained after applying the fuzzy $c$-means algorithm \cite{Bezdek1984} on the data, with $c$ being the number of decision classes. Equation \eqref{eq:fpc} formalizes this coefficient, which is intended to measure the amount of overlap between the fuzzy clusters,

\begin{equation}
\label{eq:fpc}
P =\frac1n\sum_{j=1}^c\sum_{i=1}^n{\left(\mu_\textit{ij}\right)}^{a}
\end{equation}

\noindent where $\mu_\textit{ij}$ represents the membership degree of the $i$-th instance to the $j$-th fuzzy cluster, while $a>1$ stands for the fuzzification parameter. A popular choice for this parameter is $a=2$, hence we will adopt this setting in the numerical simulations performed in this paper.

Other meta-features concern the number of presumably correct instances, which share the same decision classes as other instances in their neighborhood. In contrast, presumably incorrect instances are labeled differently when compared with their neighbors, even when these instances are strong members of their neighborhoods. Equations \eqref{eq:correct} and \eqref{eq:incorrect} show the presumably correct and incorrect decision sets for the $k$-th decision class,

\begin{equation}
\label{eq:correct}
C_l = \{ x \in X : f(x) = y_l \land g(x)=y_l \}
\end{equation}
\begin{equation}
\label{eq:incorrect}
I_l = \{ x \in X : f(x) = y_l \land g(x) \neq y_l \}
\end{equation}

\noindent where $X$ is the dataset being processed, $y_l$ is the $l$-th decision class, $f(x)$ is the ground-truth decision class, $g(x)$ is the most popular decision class in the instance's neighborhood. In this paper, the neighborhood is defined by the $k$ closest neighbors, where $k$ is the number of decision classes. Since this measure is class-dependent, we can compute the minimum, maximum, mean and standard deviation across decision classes. These meta-features are denotes as \texttt{min\_presum\_correct}, \texttt{max\_presum\_correct}, \texttt{mean\_presum\_correct} and \texttt{std\_presum\_correct}, respectively.


To calculate an anomaly score for each sample of the generated datasets, we can apply local outlier detection based on the local density deviation. The negative outlier factor (\texttt{neg\_outlier\_factor}) determines how close the sample is to the local density with the given k-nearest neighbors \cite{Cheng2019}. By comparing a sample's local density to its neighbors' local densities, we can identify samples with a notably lower density than their neighbors.

It is important to inspect how the variables in the dataset are spread around the mean. Therefore, the variance of the problem variables is integrated into the meta-dataset as a meta-feature. For this measure, we retrieve the minimum (\texttt{min\_variance}), maximum (\texttt{max\_variance}), average (\texttt{mean\_variance}), and standard deviation (and \texttt{std\_variance}) values.


Positive and negative covariance values are calculated to examine the joint variability of two features. Given that this measure can be computed for pairs of features, we calculate the minimum negative covariance (\texttt{max\_neg\_cov}) and the maximum positive covariance (\texttt{max\_pos\_cov}). In addition, the mean and standard deviation of both positive and negative covariance values are retrieved. These meta-features are represented as the following, \texttt{mean\_pos\_cov}, \texttt{mean\_neg\_cov}, \texttt{std\_pos\_cov}, \texttt{std\_neg\_cov}.


To characterize how the samples negatively and positively tailed to the normal distribution \cite{vanschoren2018meta}, we use the skewness and kurtosis. To increase the expressiveness of the meta-features, we compute the minimum, maximum, average and standard deviation values of both positively and negatively tailed samples. These meta-features are represented as \texttt{max\_neg\_skew}, \texttt{std\_neg\_skew}, \texttt{mean\_neg\_skew}, \texttt{max\_pos\_skew}, \texttt{std\_pos\_skew}, \texttt{mean\_pos\_skew}, \texttt{max\_neg\_kurtosis}, \texttt{std\_neg\_kurtosis}, \texttt{mean\_neg\_kurtosis}, \texttt{max\_pos\_kurtosis}, \texttt{std\_pos\_kurtosis}, \texttt{mean\_pos\_kurtosis}, respectively.

Another measure related to the distribution of the instances in the dataset is the index of \texttt{dispersion} \cite{ALI2006119}. In this measure the larger the values, the more scattered the data points. Thus, smaller values describe a dataset with clustered data. Equation \eqref{eq:dispersion} shows how to compute this measure,

\begin{equation}
\label{eq:dispersion}
D = \frac{kurt(n^2 - \sum_l p_l^2)}{n^2 (c-1)}
\end{equation}

\noindent where $kurt(.)$ represents the kurtosis function, $n$ is the number of instances, $c$ denotes the number of decision classes and $\sum_l p_l^2$ is the sum of the squared frequencies of the classes. 

Since features contribute independent bits of useful information, we can estimate how many features would be required to solve the problem. This can be done by computing the ratio between the class entropy and the average mutual information. This measure is referred to as the number of equivalent features (\texttt{n\_equiv\_features}) and computed as follows:

\begin{equation}
\label{eq:n_equiv_features}
Q = E(y)/ \left( \frac{1}{|X|^{-1}} \sum_{x \in X} M(x,y) \right).
\end{equation}

Another measure to compute the association between a given feature and the decision classes refers to the uncertainty coefficient, 

\begin{equation}
\label{eq:uncertainty}
U(x) = M(x,y) / E(y).
\end{equation}

Since the uncertainty coefficient is computed for each feature-target pair, we can report the minimum, maximum, mean and standard deviation, which are denoted as \texttt{min\_uncertainty}, \texttt{max\_uncertainty}, \texttt{mean\_uncertainty} and \texttt{std\_uncertainty}, respectively.


The following landmarking meta-features involve training weaker learners. For example, we can benefit from the 1-nearest neighbors  \cite{Pfahringer2000} to examine the data sparsity. Linear discriminant analysis is used to create relations between the features by retrieving whether the linear separations are present in the data \cite{vanschoren2018meta}. Additionally, a Gaussian Naive Bayes classifier helps capture the independence among the generated features \cite{vanschoren2018meta}. Calculating the average scores using cross-validation is important since the models are highly susceptible to overfitting. These meta-features are represented as \texttt{1nn\_mean\_acc}, \texttt{lda\_mean\_acc}, \texttt{nb\_mean\_acc}, respectively. Lastly, a decision tree is cross-validated for each dataset. The average performance score of this rule-based classifier (\texttt{dt\_mean\_acc}) gives information on the division of the decision space using the most informative features \cite{Pfahringer2000}. Concerning the tree structure, we retrieve other meta-features such as the number of leaves (\texttt{dt\_leaves}), the depth of the tree (\texttt{dt\_depth}), and the Gini importance. Similarly to other multi-valued metrics, we get the maximum, minimum, mean, and standard deviation Gini importance values. They are denoted as \texttt{max\_gini\_importance}, \texttt{min\_gini\_importance}, \texttt{mean\_gini\_importance}, \texttt{std\_gini\_importance}, respectively, thus completing our compact set of 62 meta-features.


\section{Approaches for synthetic data generation}
\label{sec:generation}

The next step of our autoML methodology consists of generating a collection of synthetic datasets with different complexity levels to create the meta-dataset. In this section, we will elaborate on two different approaches where the synthetic classification problems are either generated from scratch or a collection of real-world problems using a generative model.

\subsection{Generating classification problems from scratch}
\label{sec:generation:synthetic}

To generate synthetic classification problems from scratch, we will resort to the \texttt{make\_classification} function from \texttt{Sklearn}. This function produces clusters of normally distributed data points around the vertices of a \texttt{n\_informative}-dimensional hypercube, such that \texttt{n\_informative} denotes the number of informative features. The sides of the hypercube have a length equal to 2*\texttt{class\_sep}, where \texttt{class\_sep} regulates the spreadness of the decision classes. Moreover, this function allows introducing interdependence between the predictive features and adding noise to the data points.

When generating the synthetic datasets, the function parameters are randomly selected as follows\footnote{If no confusion arises, we will use $x \sim U_{\mathbb{R}}(a,b)$ and $y \sim U_{\mathbb{N}}(a,b)$ to denote randomly generated real and integer values in the $[a,b]$ interval, respectively.}. The number of samples $\texttt{n\_samples} \sim U_{\mathbb{N}}(500,50000)$ and the number of decision classes $\texttt{n\_classes} \sim U_{\mathbb{N}}(2,10)$. The number of features \texttt{n\_features} = $h \times \texttt{n\_classes}$, where $h \sim U_{\mathbb{N}}(5,20)$. Moreover, the number of informative features \texttt{n\_informative} = $\lfloor n_i \times \texttt{n\_features} \rfloor$ where $n_i \sim U_{\mathbb{R}}(0.2,0.4)$, whereas the number of redundant features \texttt{n\_redundant} = $\lfloor n_r \times (\texttt{n\_features}-\texttt{n\_informative}) \rfloor$ where $n_r \sim U_{\mathbb{R}}(0.2,0.4)$. Similarly, the number of repeated features \texttt{n\_repeated} = $\lfloor n_p \times (\texttt{n\_features} - \texttt{n\_informative} - \texttt{n\_redundant}) \rfloor$ where $n_p \sim U_{\mathbb{R}}(0.2,0.4)$. The number of clusters per class $\texttt{n\_clusters\_per\_class} \sim U_{\mathbb{N}}(1,5)$ whereas the proportions of samples assigned to each decision class is \texttt{weights}= $U_{\mathbb{R}}(0.4,1.0)$. It must be fulfilled that $\texttt{n\_clusters\_per\_class} < \lfloor 2^q / \texttt{n\_classes} \rfloor$ such that $q=\texttt{n\_informative}$. The \texttt{flip\_y} parameter is selected from the set $\{0.01, 0.05\}$ and denotes the fraction of samples whose class is assigned randomly. Larger values introduce noise in the labels and make the classification task harder. Finally, the \texttt{hypercube} parameter takes random boolean values. If true, the clusters are allocated at the vertices of a hypercube. If false, the clusters are allocated at the vertices of a random polytope. Finally, \texttt{class\_sep} $\sim U_{\mathbb{N}}(1,5)$ is a factor multiplying the hypercube size where larger values spread out the clusters (denoting the decision classes), making the classification task easier.

\subsection{Generating classification problems from real datasets}
\label{sec:generation:real}

To generate synthetic classification problems from a collection of real-world datasets, we will resort to a generative model, Conditional Tabular GAN (CTGAN), which is introduced and is shown to outperform GAN-based or Bayesian network-based data generation models in \cite{xu2019modeling}. 
The structural properties of tabular data make realistic data generation challenging for standard GAN approaches. These challenges include having mixed data types (continuous and discrete), having a non-Gaussian distribution of the continuous data (unlike image pixel data, columns having multiple modes) and discrete columns being imbalanced. These imply the GAN methods being outperformed by Bayesian network methods in realistic synthetic data generation over ``likelihood fitness"  and ``machine learning efficacy". Likelihood fitness stands for columns in the synthetic data following the same joint distribution as the real data. Machine learning efficacy is about the learned models trained on the real data (training data) performing similarly (with the test data) on the synthetic data. The conditional generator embedded in the CTGAN model overcomes these challenges. The name CTGAN relates to the way this method handles categorical data generation by conditional probability distributions. Discrete valued features can be represented as one-hot-encoded vectors. For a multi-modal continuous variable, using a min-max normalization may fail to capture the complexity of the distribution. The mode-specific normalization used in \cite{xu2019modeling} involves calculating the probability of each mode, coming up with a corresponding vector representation, and concatenating these vectors as well as the one-hot-encoded vectors for discrete features to represent a row in the data. 


\section{The proposed model selection approach}
\label{sec:proposal}

Before presenting our methodology, we should define the \textit{multilabel model selection problem}. Let $\Omega = \{\Omega_1, \dots, \Omega_k\}$ be a set of classification algorithms with hyperparameters $\Delta = \{\Delta_1, \dots, $ $ \Delta_k\}$ such that $\Delta_i$ is the set of hyperparameters for the $i$-th classifier. Moreover, let us define $\delta_{i}^{j} \in \Delta_i$ as the $j$-th hyperparameter and $\mathcal{D}(\delta_{i}^{j})$ as its domain. The set of all settings for the $i$-th classifier is $\Pi_{i} = \bigtimes_j \mathcal{D}(\delta_{i}^{j})$ while the set of all models is given by $\Pi = \bigcup_i \Omega_i \bigtimes \Pi_{i}$. Observe that a model can be understood as a classifier using a specific hyperparameter setting. Therefore, the multilabel model selection problem consists of predicting which models will yield the best performance possible w.r.t. a given metric for an unseen classification dataset. Finally, the \textit{model selection problem} consists of selecting $\pi_i \in \Pi $ with the highest likelihood of having the highest performance among all explored models.

\subsection{Building the meta-dataset with multilabel instances}
\label{sec:proposal:metadataset}

This sub-section explains the process of creating the meta-datasets relating the meta-features describing the statistical properties of each problem with the models' performance. The labels can be obtained from the set of all models is given by $\Pi = \bigcup_i \Omega_i \bigtimes \Pi_{i}$, such that $p=|\Pi|$ denotes the total number of labels. This procedure is gathered into two well-defined steps. Firstly, each synthetic dataset is split into two disjoint sets, namely, the training set (80\%) and validation set (20\%) such that each model $\pi_j \in \Pi$ is fit on the training data and evaluated on the validation set, thus producing an error vector. Secondly, the best-performing models (determined using a performance threshold) are positively labeled, while those that do not reach the optimal performance are negatively labeled. These positive and negative labels can be encoded as a binary vector, representing the output for the meta-feature vector describing the classification problem being encoded.

\begin{figure}[!ht]
     \centering
     \begin{subfigure}{0.4\textwidth}
         \centering
         \includegraphics[width=\textwidth]{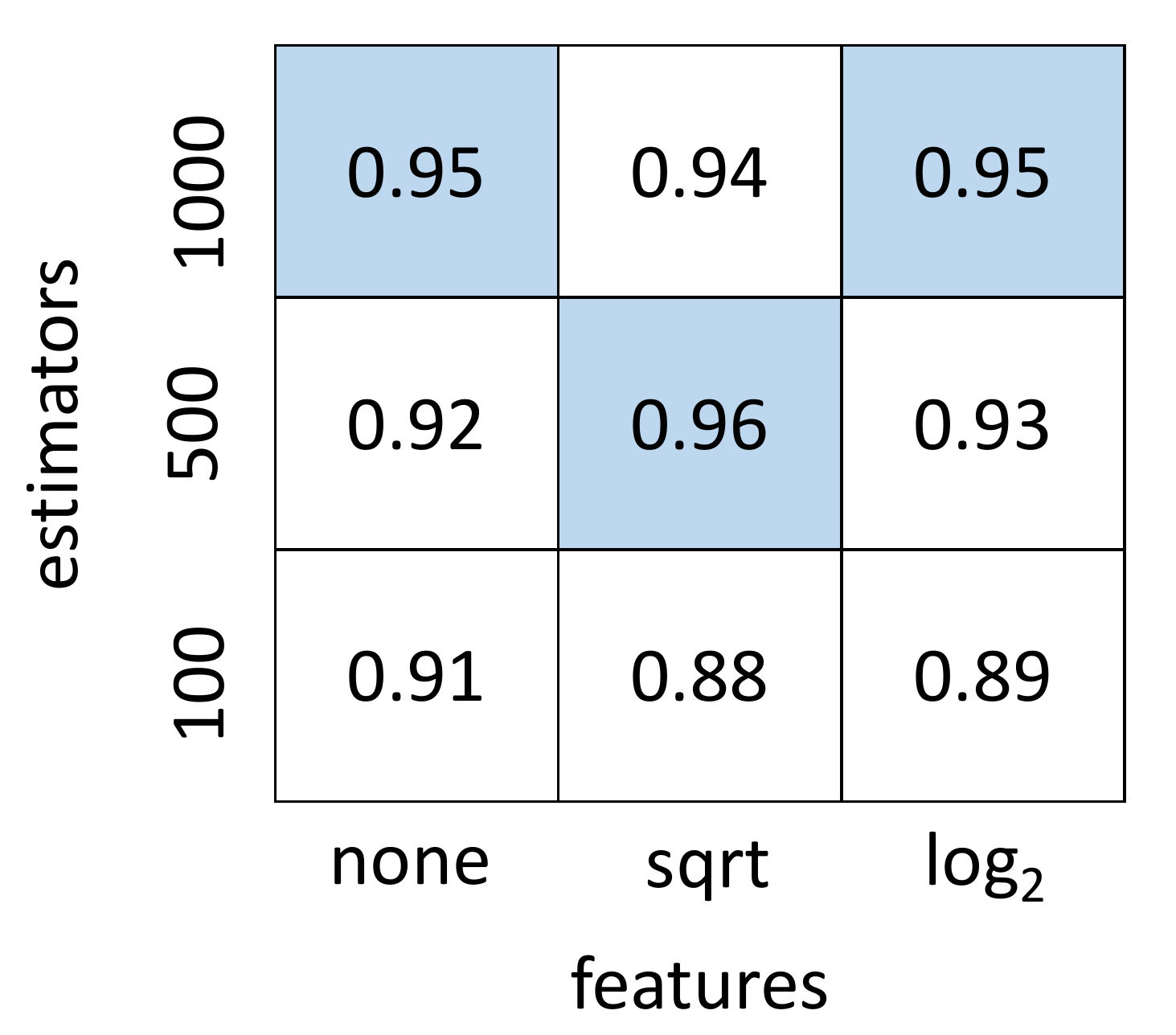}
         \caption{RF}
     \end{subfigure}
     \begin{subfigure}{0.40\textwidth}
         \centering
         \includegraphics[width=\textwidth]{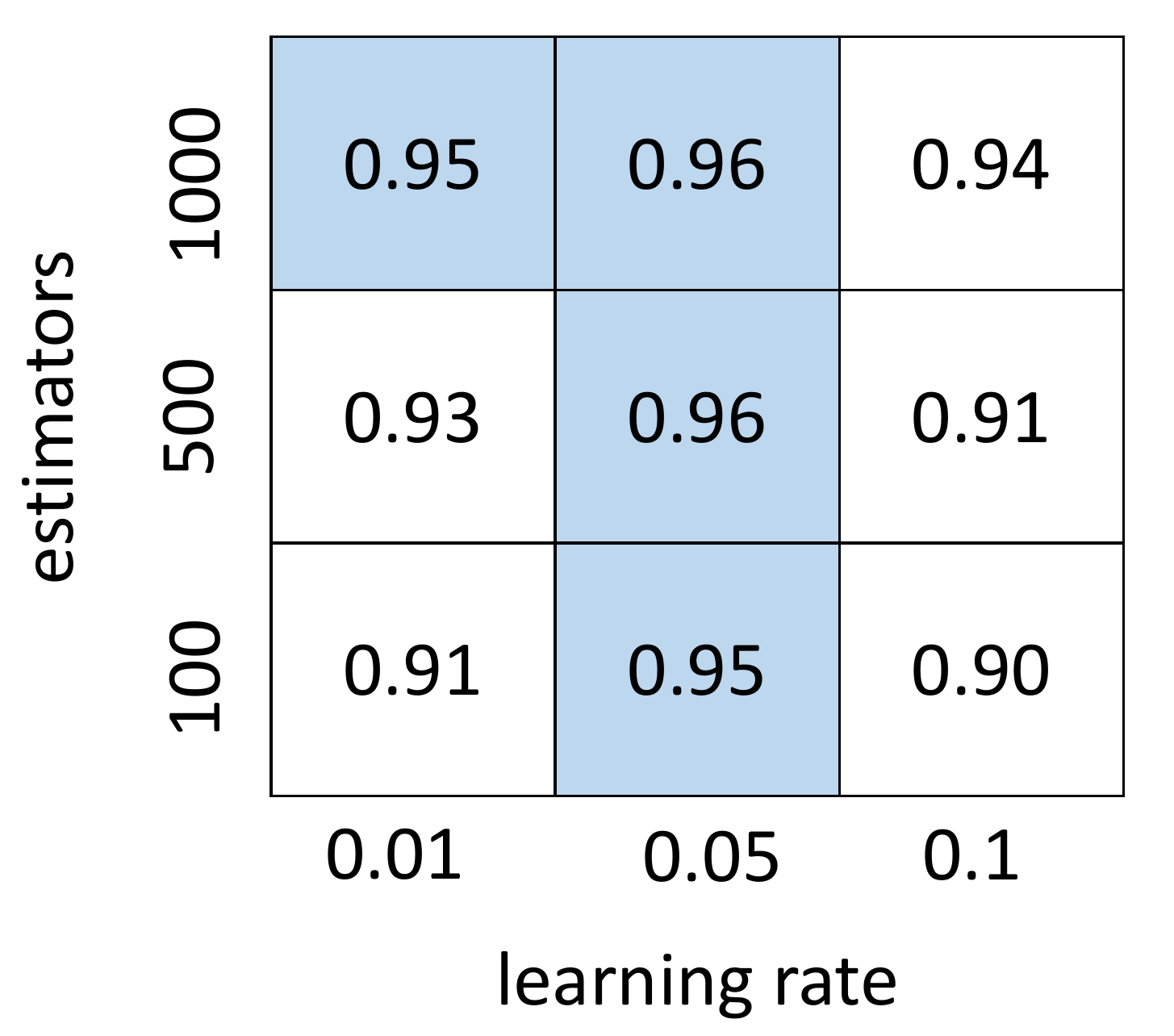}
         \caption{LGBM}
     \end{subfigure}
        \caption{Hyperparameter accuracy grid for (a) Random Forest and (b) Light Gradient Boosting Machine. In this example, each algorithm optimizes two hyperparameters over three possible values. The models highlighted in blue will be positively labeled since their accuracy and best accuracy scores differ by 1\% at most in both performance grids.}
    \label{fig:hp-configurations}
\end{figure}

Figure \ref{fig:hp-configurations} shows, as an example, two-parameter grids concerning Random Forest (RF) and Light Gradient Boosting Machine (LGBM) using accuracy as the performance metric. Each algorithm optimizes two hyperparameters (the number of estimators and features in the case of RF and the number of estimators and learning rate in the case of LGBM) over three possible values, totaling 18 configurations associated with the given dataset. The models highlighted in blue can be positively labeled since their accuracy scores and best accuracy value differ by 1\% at most in both grids. In contrast, the remaining models will be negatively labeled. To obtain the label vector, the resulting binary masks are flattened and merged into a single binary vector having 18 dimensions encoding 7 positive and 11 negative models.

\subsection{CNN-based meta-classifier for model selection}
\label{sec:proposal:cnn}

Aiming to tackle the model selection problem, we need a meta-classifier mapping the meta-feature space to the model performance space. In this paper, we use a CNN-based architecture for this task where the input is a numerical vector encoding the dataset and the output is the associated binary vector. However, CNNs are not entirely suited for handling data with no topological organization, as happens in tabular pattern classification problems. In other words, applying convolutional and pooling operators on a tabular dataset makes little sense since rows and columns are arbitrarily placed \cite{Bello2020}. One simple yet effective way to overcome this issue is by feeding the CNN with a set of non-linear combinations of problem features involving learnable parameters. In practice, this is materialized by adding a fully connected layer after the input layer and before any convolution or pooling layer.

\begin{figure}[!ht]
\centering
  \includegraphics[width=\textwidth]{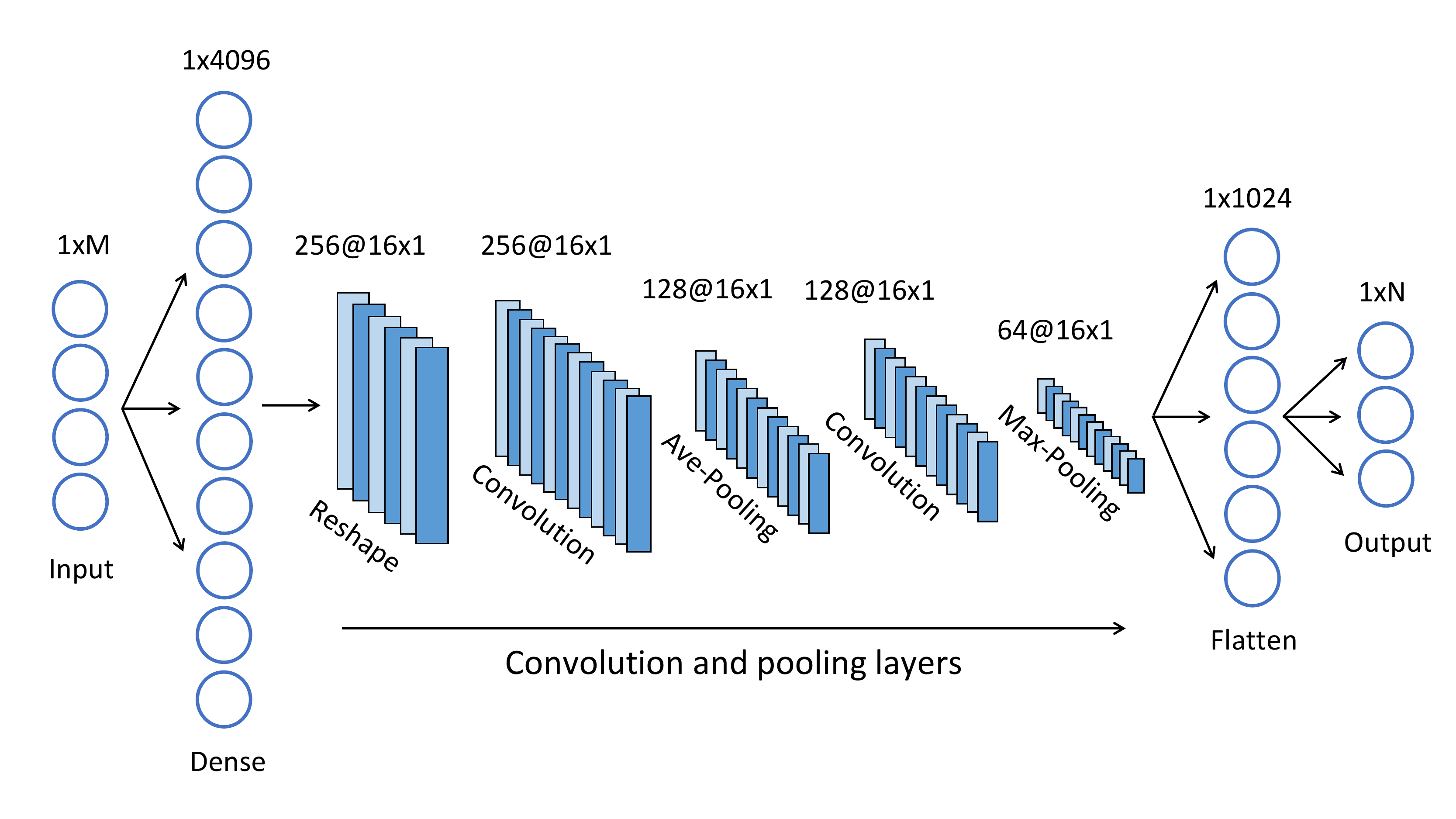}
\caption{CNN model for model selection. This network receives a meta-feature vector encoding a classification problem and produces the algorithms and parameter settings likely to reach optimal performance.}
\label{fig:cnn-model}
\end{figure}

The proposed CNN architecture starts with the mentioned fully-connected layer having 4096 hidden neurons followed by a dropout layer. After a reshaping operation that generates a $256 \times 16$ matrix, we add a batch normalization layer to speed up the convergence and reduce the number of training epochs. In short, this layer applies a transformation that maintains its mean close to zero and its standard deviation close to one. This is followed by another dropout layer and a convolutional layer having 16 filters and a kernel size of 5. Subsequently, we add an average pooling layer with a pooling size of 2, followed by a batch normalization layer and a dropout layer. The following layers in this deep neural network are a convolutional layer having 16 filters and a kernel size of 3, and a maximum pooling layer with a pooling window of 4 and two strides. Finally, we perform a flattening operation followed by an output layer having as many neurons as classification models (labels). Figure \ref{fig:cnn-model} shows a summary of the CNN model used as a meta-classifier that excludes the batch normalization and dropout layers.

In this model, we use the squared hinge function as the activation function in the last layer as it allows maximizing the margins between optimal and non-optimal models \cite{Napoles2011}. We say that a candidate classification model is positive if the corresponding output neuron produces a positive value; otherwise, the model will be deemed negative. Moreover, we will use the binary cross-entropy as the loss function and the ADAM optimizer. As for the hyperparameters to be optimized, we will fine-tune the activation function used in the inner layers, the dropout rate, and the learning rate.

\section{Numerical simulations}
\label{sec:simulations}

In this section, we evaluate the performance of the proposed autoML approach using different sets of meta-features and meta-classifiers. Moreover, we use entirely synthetic or partially synthetic datasets generated from real-world classification problems. If no confusion arises, we will refer to the former as synthetic datasets and the latter as real datasets.

\subsection{Algorithms and hyperparameters}
\label{sec:simulations:algorithms}

In the experiments, the classifiers to be recommended concern Random Forest (RF) and Light Gradient Boosting Machine (LGBM) as they are proven to be strong learners. In the case of RF, the hyperparameters to be optimized are the information criterion (\texttt{gini} or \texttt{entropy}), the strategy for selecting the maximum number of features (\texttt{sqrt}, \texttt{log2} or \texttt{None}) and the number of estimators (500 or 1000). In the case of LGBM, the hyperparameters to be optimized are boosting type (\texttt{gbdt} or \texttt{dart}), the learning rate (0.1, 0.05, 0.01) and the number of estimators (500 or 1000). As a result, each ensemble learner is associated with 12 configurations, thus resulting in 24 models.

As for the meta-classifiers, we compare the proposed CNN-based model with the following multilabel algorithms: Binary Relevance (BIREL) \cite{godbole2004discriminative,zhang2018binary}, RAndom k-labELsets (RAkEL) \cite{Tsoumakas2011,padmashani2019rakel}, and Multi-label kNN (MLkNN) \cite{zhang2007ml,zhu2020}, which are available in the Scikit-multilearn library \cite{szymal2019scikit}). The first and second methods implement problem transformation strategies, while the latter uses an algorithm transformation approach. 

BIREL generates a binary dataset per label such that positive patterns are associated with the label. When a new pattern is presented to the model, the output will be the set of positive classes. RAkEL generates random subsets of labels while training a multi-label classifier for each subset. MLkNN is an adaptation of $k$NN to the multi-label classification scenario. This lazy learner finds the nearest instance to a test class and uses Bayesian inference to select assigned labels of unseen instances. 

During nested cross-validation (using grid search), BIREL and RAkEL use a decision tree as the base classifier. Moreover, we will optimize the information criterion (\texttt{gini} or \texttt{entropy}) and the strategy for selecting the maximum number of features (\texttt{sqrt}, \texttt{log2}). The hyperparameter $k$ of MLkNN represents the number of neighbors (3, 5, 7 or 10). In the CNN model, the hyperparameters to be optimized are the learning rate (0.001, 0.005, 0.01), the dropout rate (from 0.0 to 0.9 with step equal to 0.1) and the neuron's activation function (\texttt{relu}, \texttt{elu}, \texttt{selu}). All simulations were performed on a high-performance computing environment that uses two Intel Xeon Gold 6152 CPUs at 2.10 GHz, each with 22 cores and 187 GB of memory. 

\subsection{Results and discussion}
\label{sec:simulations:discussion}

To assess the meta-classifiers' predictions, we will use several evaluation metrics. Within the multilabel metrics, we will compute the Hamming loss and the macro versions of precision, recall, specificity and F1-score, where we first calculate the metric for each label and then report the unweighted mean. However, due to its practical relevance, the most important measure in our study is the \textit{hit rate}. It gives the proportion of instances to which the most likely positive model predicted by the meta-classifier is indeed an optimal model. Equation \eqref{eq:hit_rate} shows this performance measure,

\begin{equation}
\label{eq:hit_rate}
H(X) = \frac{1}{|X|} \sum_{x \in X}  \left\{
\begin{array}{ll}
      0, & ~\text{if}~ y_{(x)}^{+} \not\subseteq Y_{(x)}^{+} \\
      1, & ~\text{if}~ y_{(x)}^{+} \subseteq Y_{(x)}^{+} \\
\end{array} 
\right. 
\end{equation}

\noindent where $X$ represents a set of instances, $y_{(x)}^{+}$ is the most probable positive model for a given instance, as predicted by the meta-classifier, whereas $Y_{(x)}^{+}$ is the set of all known positive models associated to that instance.

In addition, we will compare our meta-features with the ones in the \texttt{pymfe} package \cite{JMLR:meta-features} in terms of prediction performance. This package allows extracting meta-features from tabular datasets containing both discrete and numerical features. Since the generated datasets only involve numerical features, we will exclude the meta-features associated with discrete ones, which can be gathered into the following categories: (1) general information measures such as the number of instances, features and decision classes, (2) standard statistical measures that describe the numerical properties of data distribution, (3) model-based measures that extract characteristics from simple machine learning models, (4) landmarking-type measures associated to the performance of simple and efficient learning algorithms, (5) clustering-based measures that extract information about dataset using external validation indexes, (6) concept-based measures that estimate the variability of class labels among problem instances, and (7) complexity-based measures that estimate the difficulty in separating the data points into their expected classes.

It should be noted that some of these measures return more than one value. However, the \texttt{pymfe} package allows aggregating these values by using a summarization strategy. In this paper, we use the mean, standard deviation, kurtosis and histograms. In addition, we filter out the low-variance meta-features and impute the missing values with a $k$-nearest neighbor imputer. After completing these steps, we end up having 350 meta-features.

Tables \ref{table:metrics:own:synthetic}, \ref{table:metrics:pymfe:synthetic}, \ref{table:metrics:own:real} and \ref{table:metrics:own:real} summarize the averaged performance metrics (after performing 5-fold cross-validation) for both datasets and two different sets of meta-features. Concerning the synthetic data, the proposed CNN model emerges as the best-performing meta-model regardless of the meta-features used to describe the classification problems. In other words, the CNN model reports the largest hit rate, precision, recall, specificity and F1-score values, which are maximization metrics, while reporting the smallest Hamming loss, which is a minimization metric. Concerning the real data, all models perform comparably in terms of hit rates when using our meta-features. However, the performance drops as soon as we use the \texttt{pymfe} meta-features, with the CNN and MLKNN being the best-performing algorithms.

\begin{table}
\centering
\caption{Performance measures of the meta-classifiers evaluated on the synthetic data using the proposed meta-features.}
\label{table:metrics:own:synthetic}
\begin{tabular}{|c|c|c|c|c|c|c|} 
\hline
Algorithm & Hit Rate & Precision & Recall & Specificity & F1-score & Hamming  \\ 
\hline
CNN       & 0.908    & 0.798     & 0.778  & 0.868       & 0.788    & 0.153    \\ 
\hline
BIREL     & 0.491    & 0.675     & 0.677  & 0.793       & 0.676    & 0.237    \\ 
\hline
RAKEL     & 0.495    & 0.670     & 0.674  & 0.790       & 0.672    & 0.240    \\ 
\hline
MLKNN     & 0.723    & 0.573     & 0.670  & 0.667       & 0.618    & 0.302    \\
\hline
\end{tabular}
\end{table}

\begin{table}
\centering
\caption{Performance measures of the meta-classifiers evaluated on the synthetic data using \texttt{pymfe} meta-features.}
\label{table:metrics:pymfe:synthetic}
\begin{tabular}{|c|c|c|c|c|c|c|} 
\hline
Algorithm & Hit Rate & Precision & Recall & Specificity & F1-score & Hamming  \\ 
\hline
CNN       & 0.902    & 0.788     & 0.779  & 0.861       & 0.783    & 0.157    \\ 
\hline
BIREL     & 0.472    & 0.666     & 0.666  & 0.790       & 0.666    & 0.243    \\ 
\hline
RAKEL     & 0.487    & 0.661     & 0.665  & 0.785       & 0.663    & 0.246    \\ 
\hline
MLKNN     & 0.799    & 0.632     & 0.707  & 0.730       & 0.667    & 0.256    \\
\hline
\end{tabular}
\end{table}

\begin{table}
\centering
\caption{Performance measures of the meta-classifiers evaluated on real datasets using the proposed meta-features.}
\label{table:metrics:own:real}
\begin{tabular}{|c|c|c|c|c|c|c|} 
\hline
Algorithm & Hit Rate & Precision & Recall & Specificity & F1-score & Hamming  \\ 
\hline
CNN       & 0.872    & 0.671     & 0.794  & 0.400       & 0.728    & 0.302    \\ 
\hline
BIREL     & 0.876    & 0.643     & 0.640  & 0.501       & 0.641    & 0.363    \\ 
\hline
RAKEL     & 0.875    & 0.646     & 0.645  & 0.500       & 0.645    & 0.360    \\ 
\hline
MLKNN     & 0.877    & 0.779     & 0.523  & 0.662       & 0.625    & 0.318    \\
\hline
\end{tabular}
\end{table}

\begin{table}
\centering
\caption{Performance measures of the meta-classifiers evaluated on real datasets using the \texttt{pymfe} meta-features.}
\label{table:metrics:pymfe:real}
\begin{tabular}{|c|c|c|c|c|c|c|} 
\hline
Algorithm & Hit Rate & Precision & Recall & Specificity & F1-score & Hamming  \\ 
\hline
CNN       & 0.828    & 0.617     & 0.720  & 0.510       & 0.664    & 0.307    \\ 
\hline
BIREL     & 0.628    & 0.551     & 0.556  & 0.584       & 0.554    & 0.379    \\ 
\hline
RAKEL     & 0.612    & 0.542     & 0.549  & 0.573       & 0.545    & 0.387    \\ 
\hline
MLKNN     & 0.833    & 0.624     & 0.560  & 0.621       & 0.589    & 0.329    \\
\hline
\end{tabular}
\end{table}

We can formalize two partial conclusions from the previous results. Firstly, the CNN model reports the largest hit rate scores compared to other meta-learners for both datasets, regardless of the adopted meta-features. Secondly, our compact set of meta-features yield better results overall. In contrast, the \texttt{pymfe} meta-features yield competitive results for the larger dataset, which makes sense if we consider that we need more instances to cover decision spaces described with more features. It should be highlighted that having fewer meta-features with high predictive power is desired as it reduces the computational complexity of building the meta-instances.

\begin{figure}
  \includegraphics[width=\textwidth]{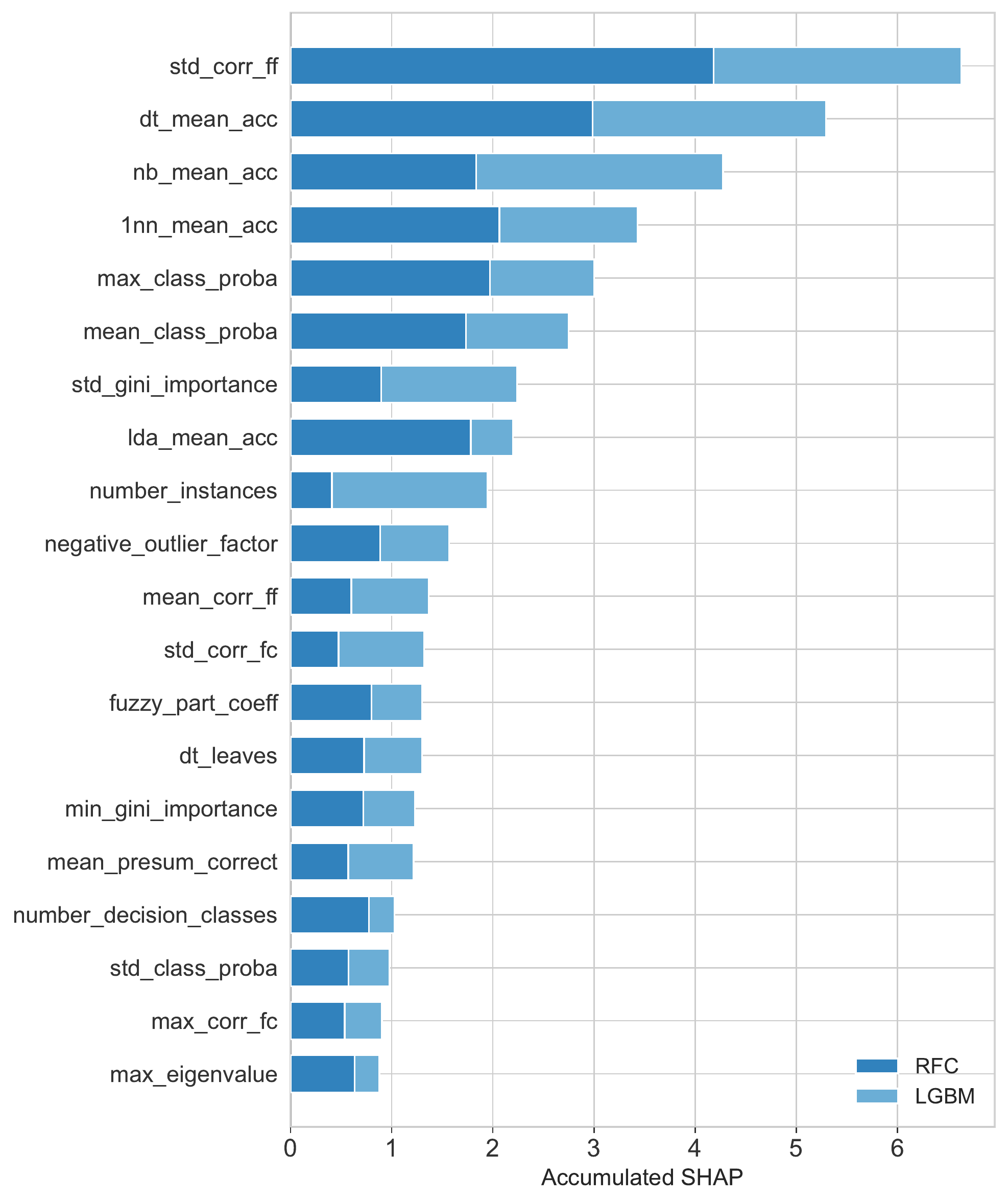}
\caption{SHAP values associated with the CNN model for the synthetic dataset. We visualize the SHAP values for both classifier families.}
\label{fig:shap:synthetic}
\end{figure}

\begin{figure}
  \includegraphics[width=\textwidth]{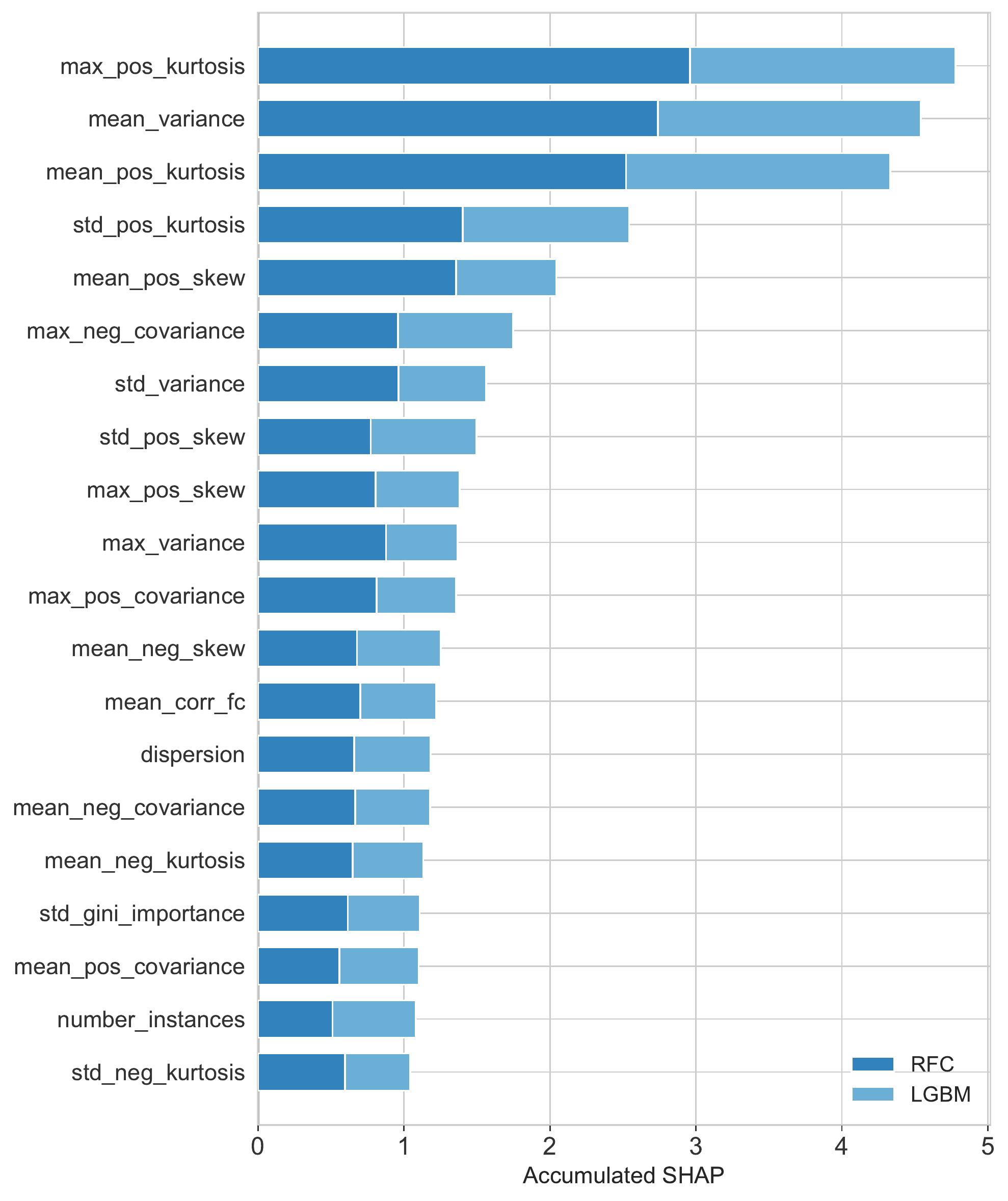}
\caption{SHAP values associated with the CNN model for the real dataset. We visualize the SHAP values for both classifier families.}
\label{fig:shape:real}
\end{figure}

Aiming to determine which meta-features lead the model selection mechanism, we resort to SHAP (SHapley Additive exPlanations) post-hoc method \cite{SHAP2017}. Since our meta-datasets are multilabel, we will aggregate all scores associated with the same classifier family. As a result, we will produce a meta-feature relevance ranking for RF models and another for LGBM models. Figure \ref{fig:shap:synthetic} and \ref{fig:shape:real} displays the relevance scores for the top-20 meta-features for the synthetic and real-world meta-datasets, respectively.

In the case of the synthetic meta-dataset, the correlation between the features, the distribution of decision classes, and the performance of weaker classifiers are reliable proxies for model selection. The novel metrics, such as the fuzzy partition coefficient and the presumably correct instances, were also included in the top-20 meta-features. In the case of the real-world meta-dataset, the measures related to the kurtosis, skewness, variance and covariance are definitely the most relevant meta-features.

\begin{figure}
     \centering
     \begin{subfigure}{0.32\textwidth}
         \centering
         \includegraphics[width=\textwidth]{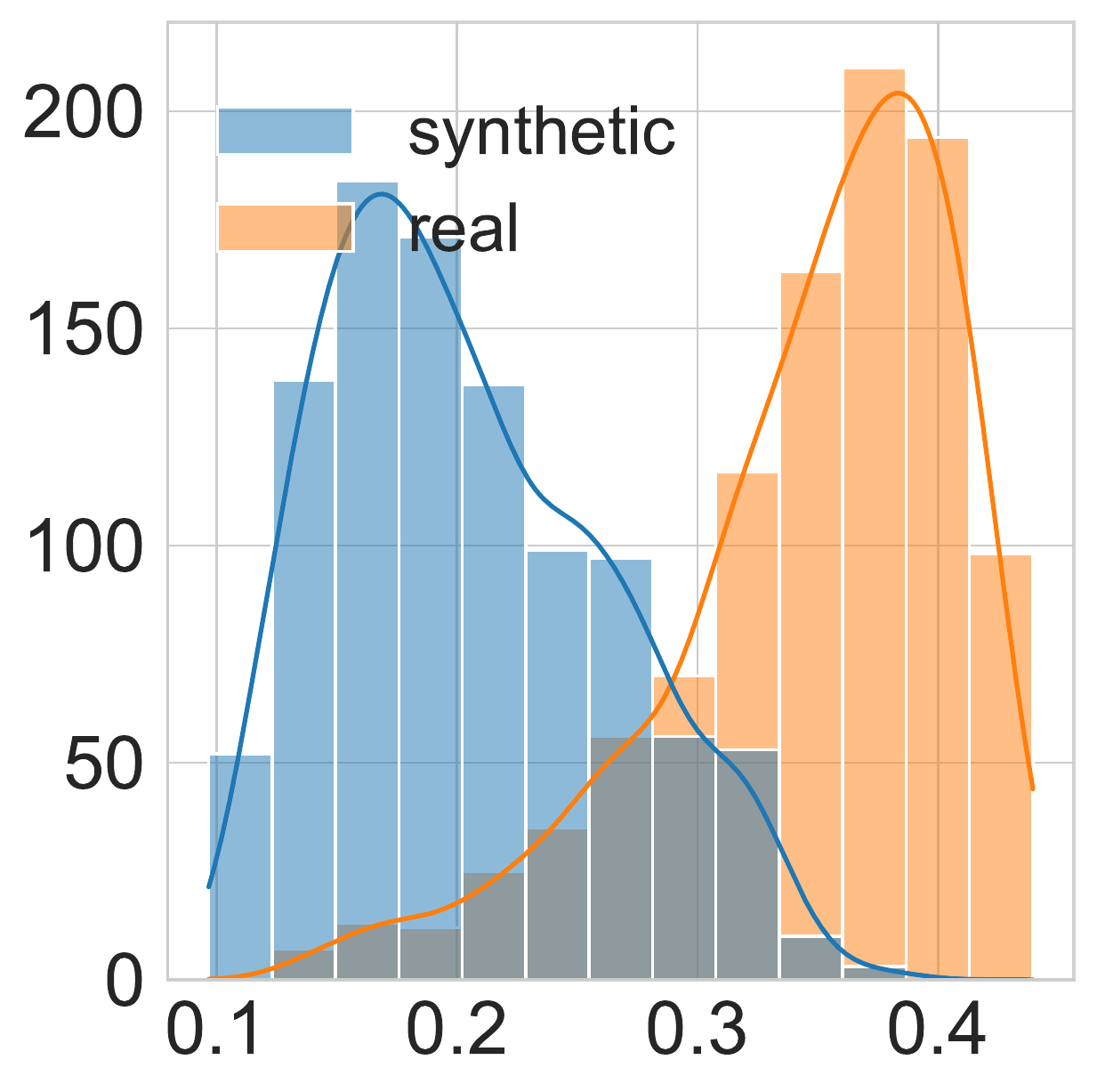}
         \caption{std\_corr\_ff}
     \end{subfigure}
     \begin{subfigure}{0.32\textwidth}
         \centering
         \includegraphics[width=\textwidth]{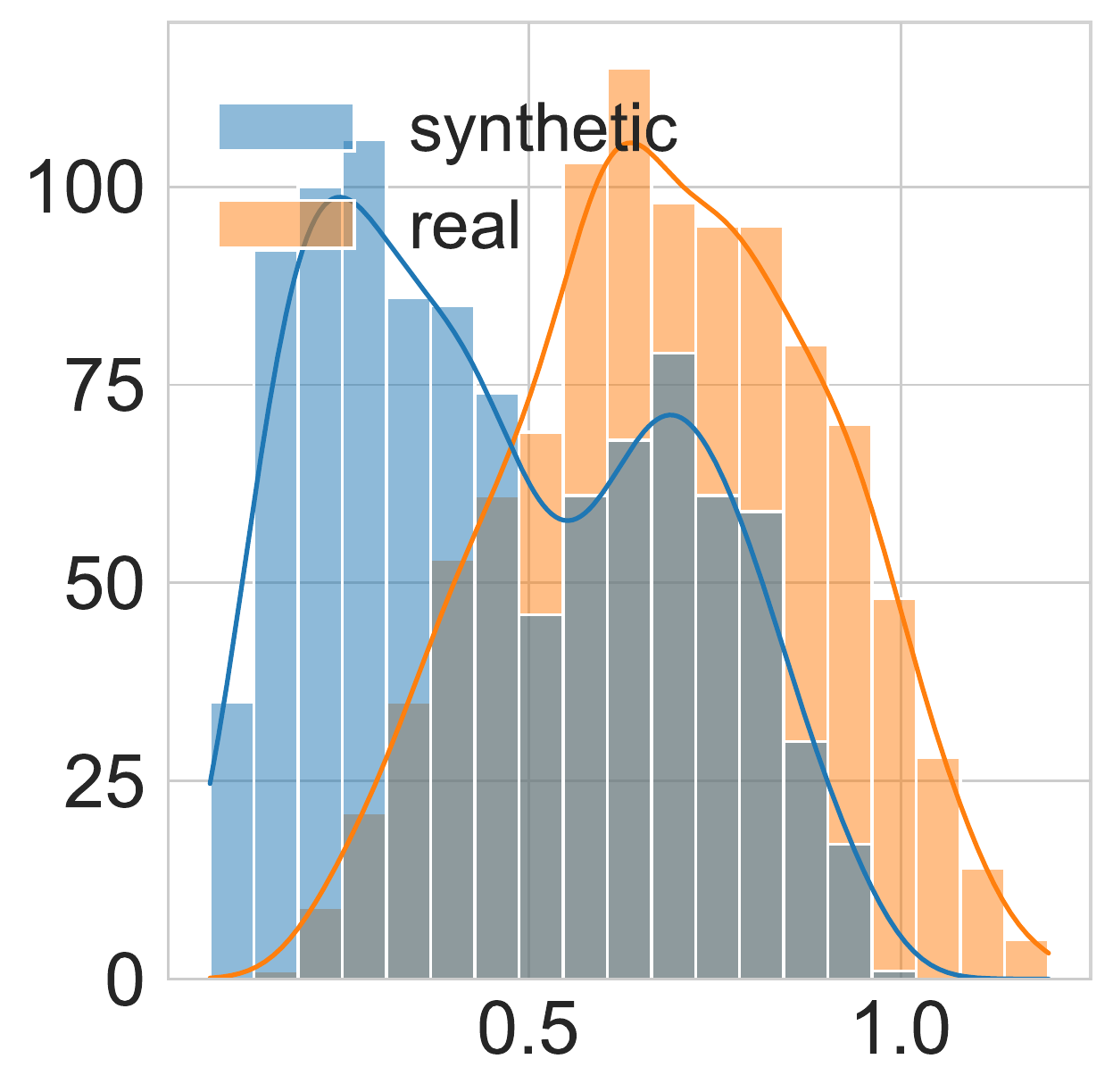}
         \caption{dt\_mean\_acc}
     \end{subfigure}
     \begin{subfigure}{0.32\textwidth}
         \centering
         \includegraphics[width=\textwidth]{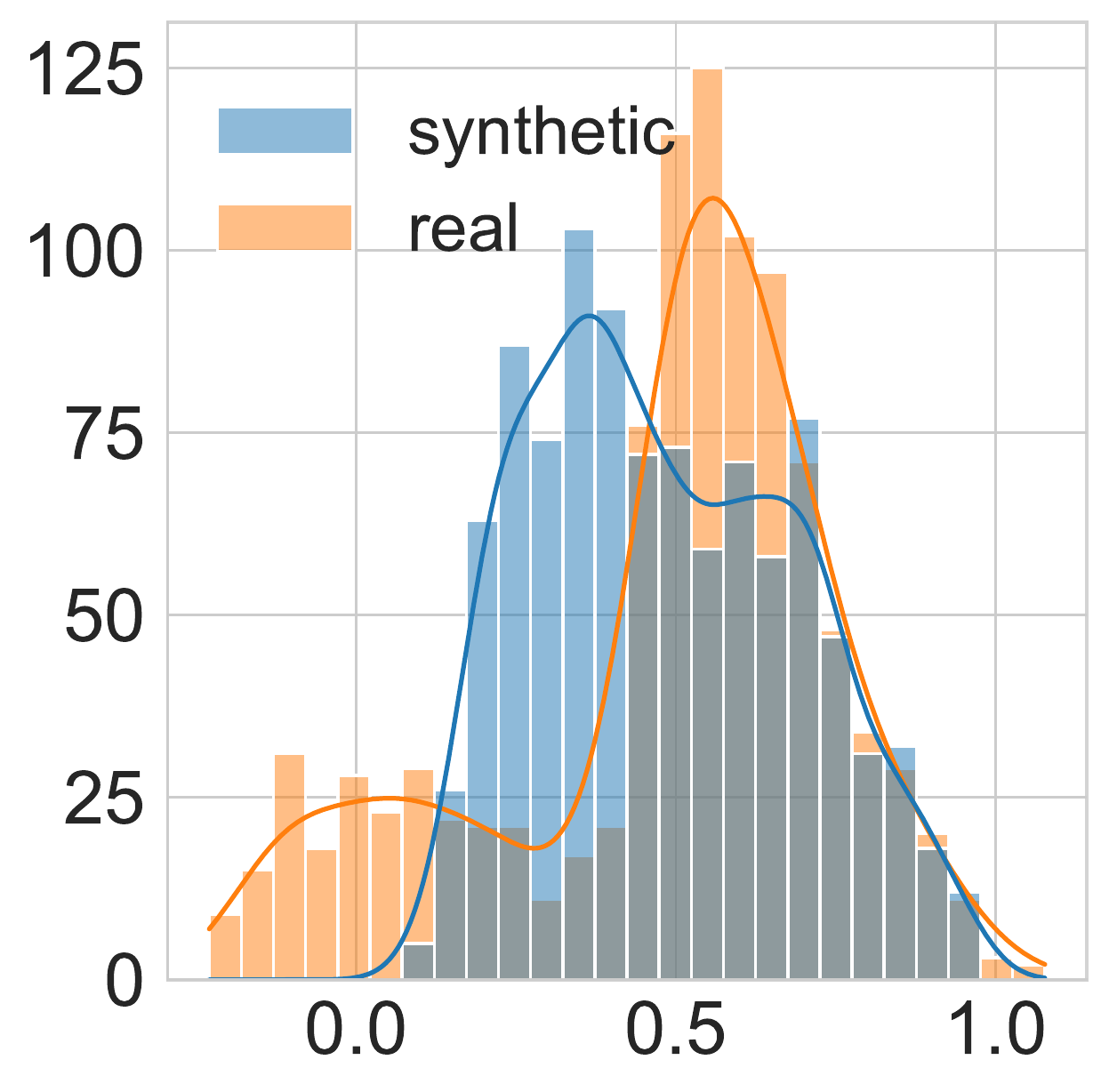}
         \caption{nb\_mean\_acc}
     \end{subfigure}

     \begin{subfigure}{0.32\textwidth}
         \centering
         \includegraphics[width=\textwidth]{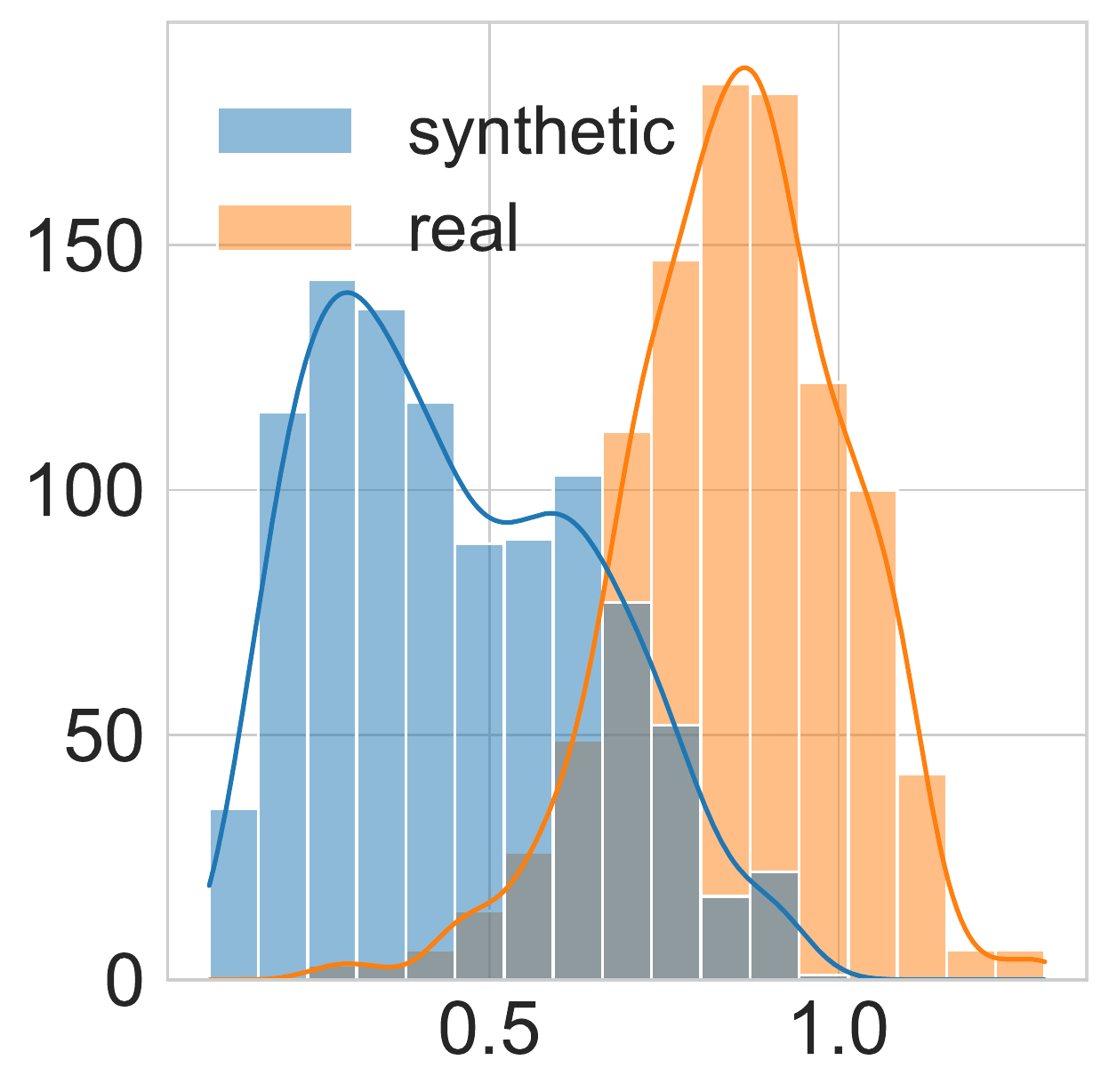}
         \caption{1nn\_mean\_acc}
     \end{subfigure}
     \begin{subfigure}{0.32\textwidth}
         \centering
         \includegraphics[width=\textwidth]{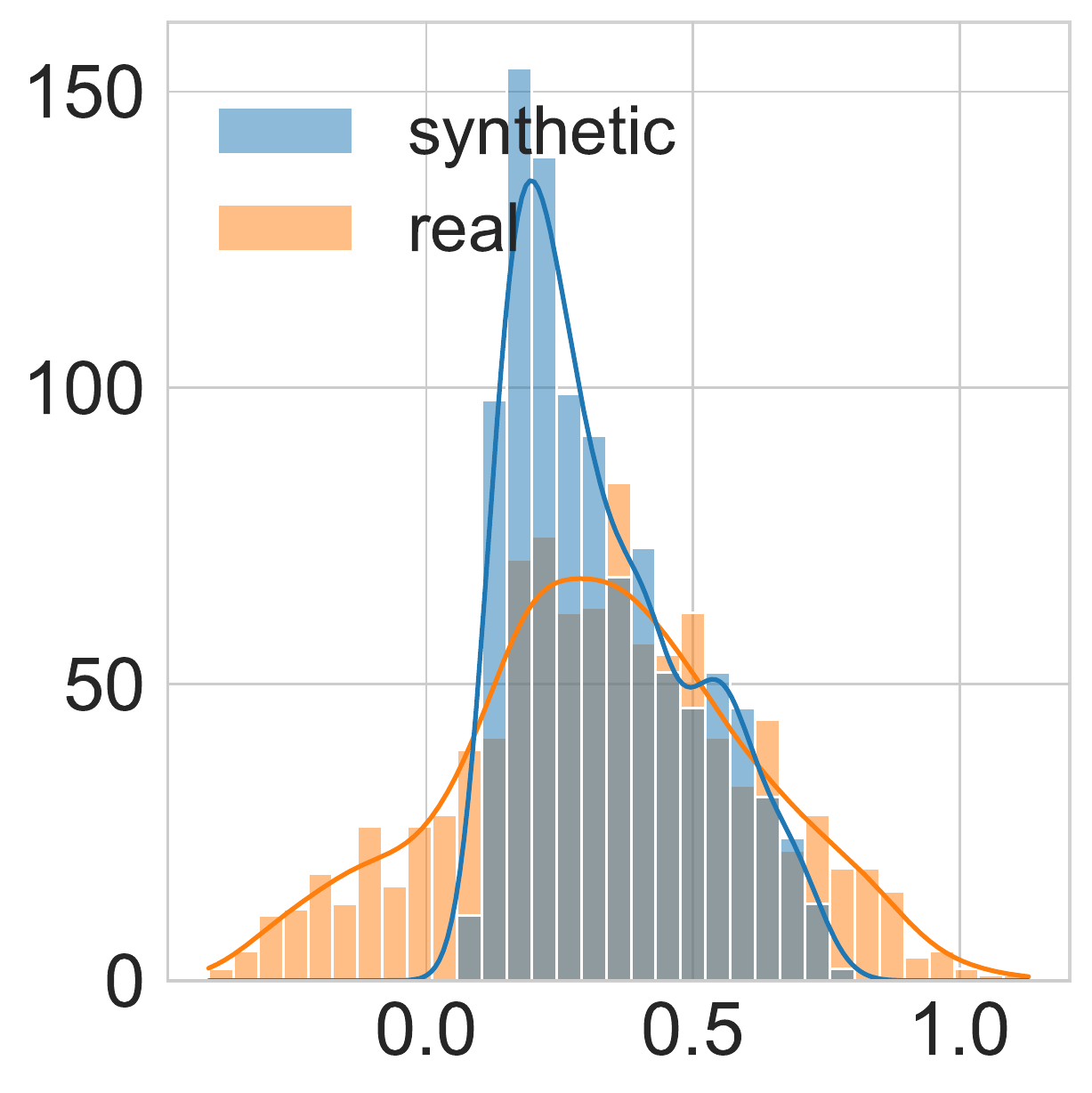}
         \caption{max\_class\_proba}
     \end{subfigure}
     \begin{subfigure}{0.32\textwidth}
         \centering
         \includegraphics[width=\textwidth]{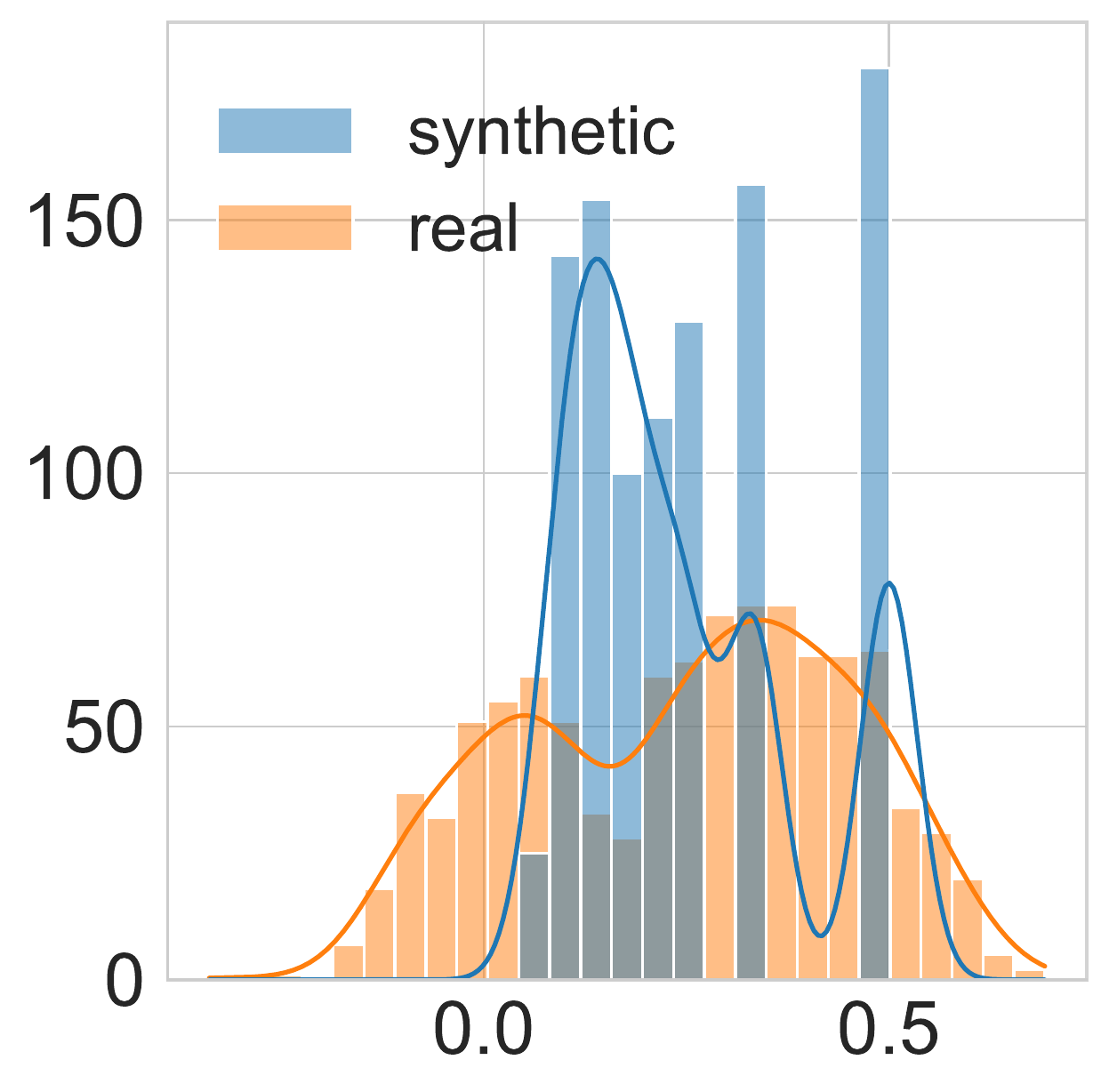}
         \caption{mean\_class\_proba}
     \end{subfigure}

     \begin{subfigure}{0.32\textwidth}
         \centering
         \includegraphics[width=\textwidth]{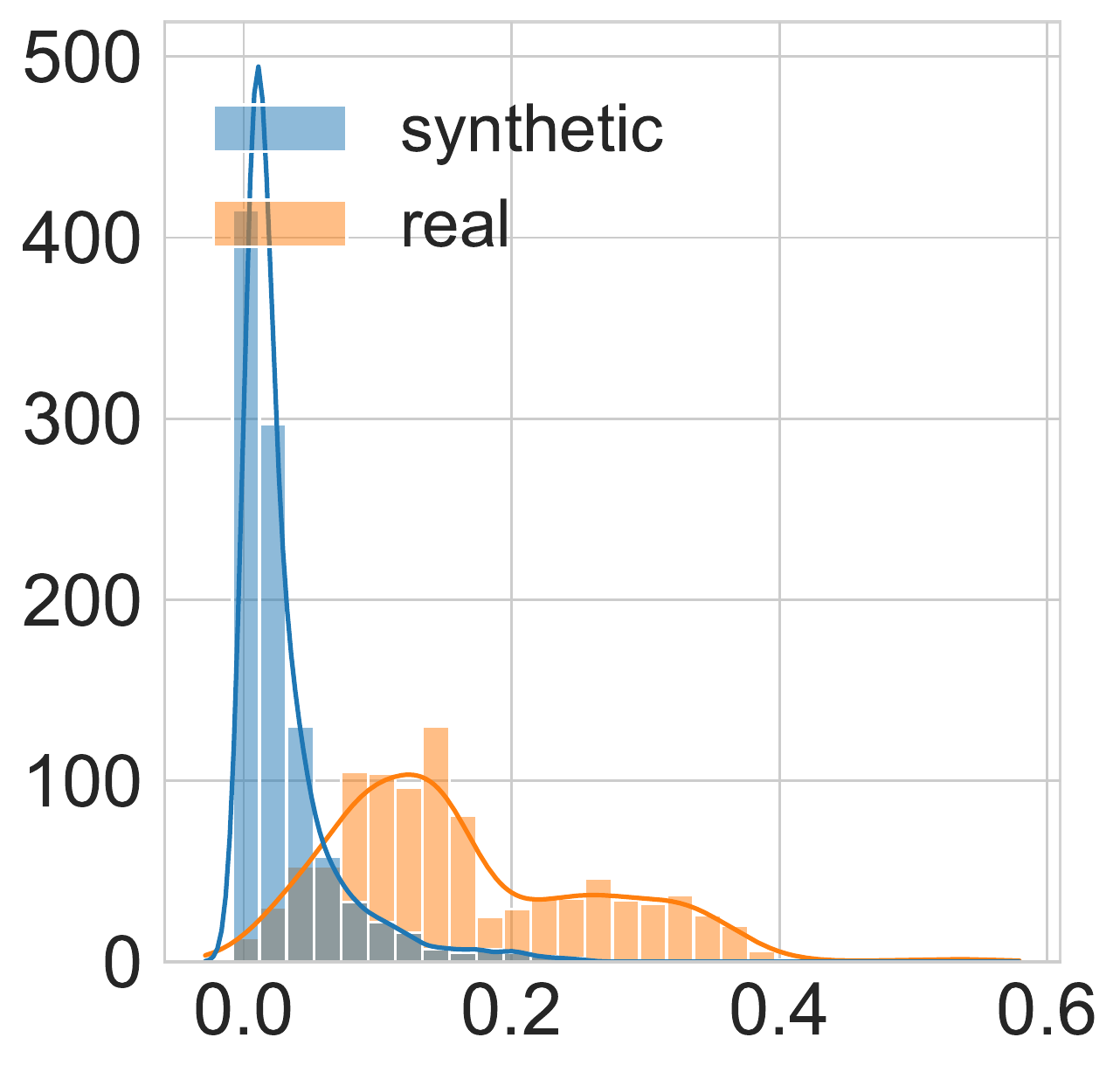}
         \caption{std\_gini\_importance}
     \end{subfigure}
     \begin{subfigure}{0.32\textwidth}
         \centering
         \includegraphics[width=\textwidth]{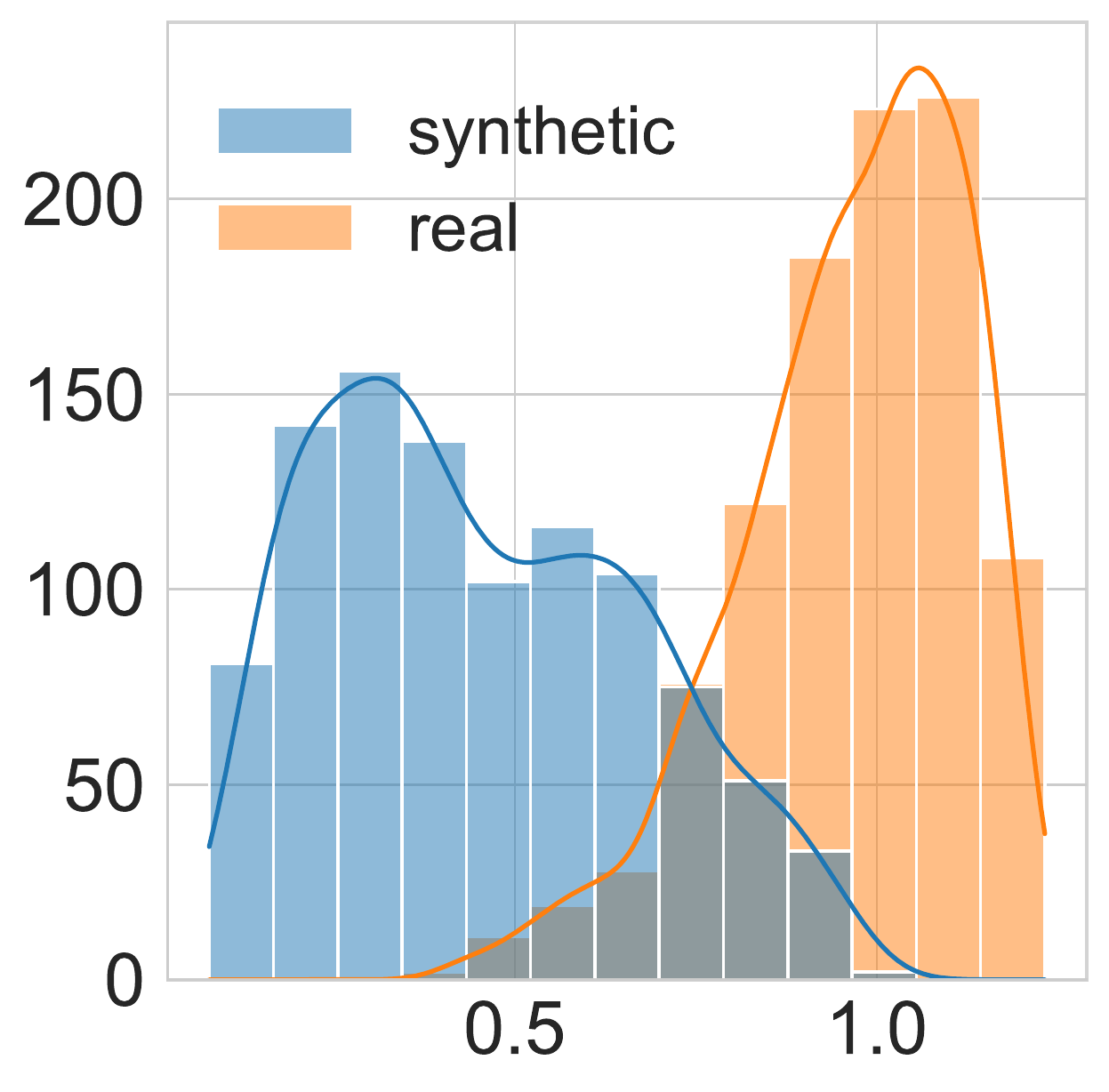}
         \caption{lda\_mean\_acc}
     \end{subfigure}
     \begin{subfigure}{0.32\textwidth}
         \centering
         \includegraphics[width=\textwidth]{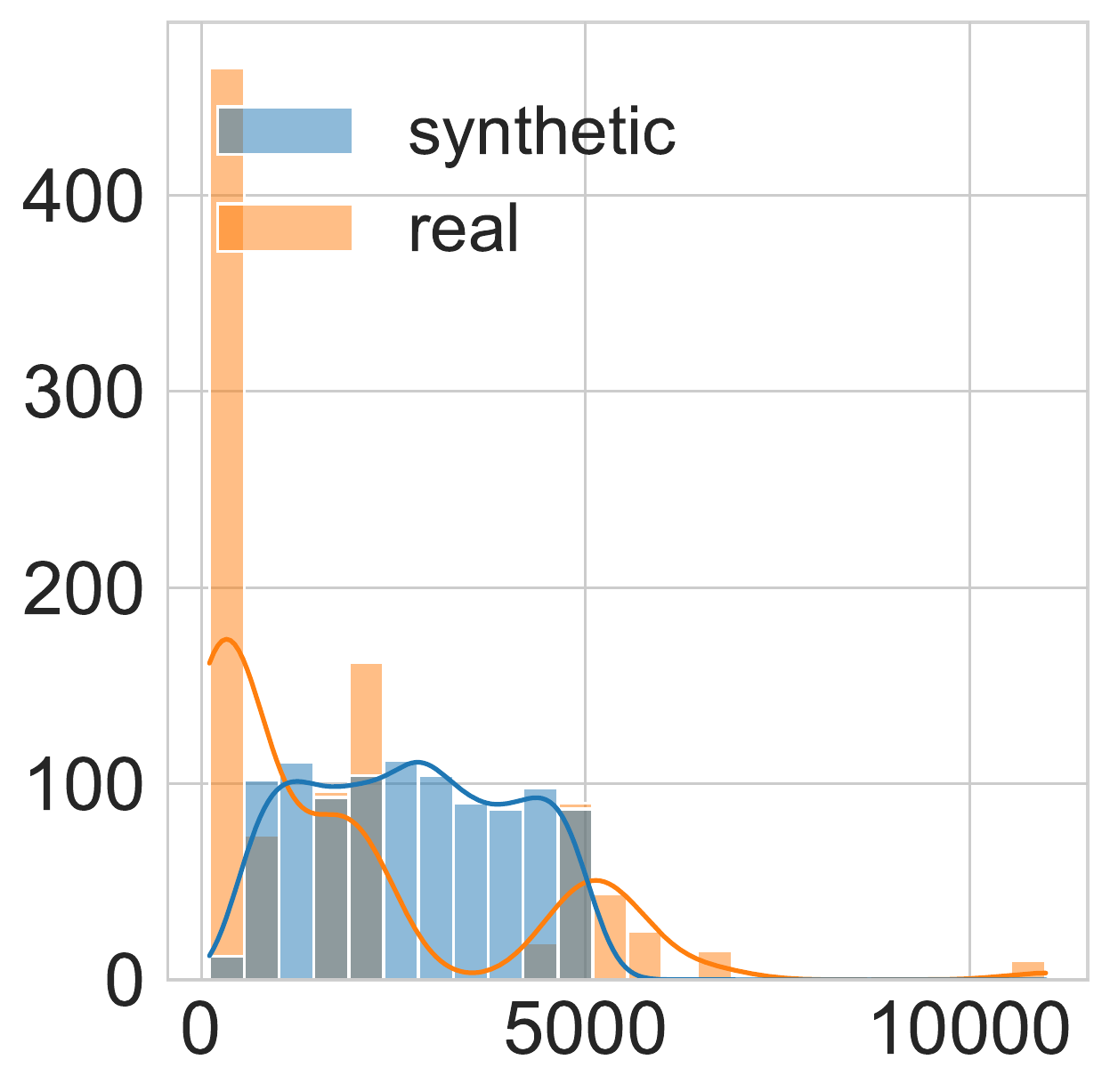}
         \caption{number\_instances}
     \end{subfigure}

     \begin{subfigure}{0.32\textwidth}
         \centering
         \includegraphics[width=\textwidth]{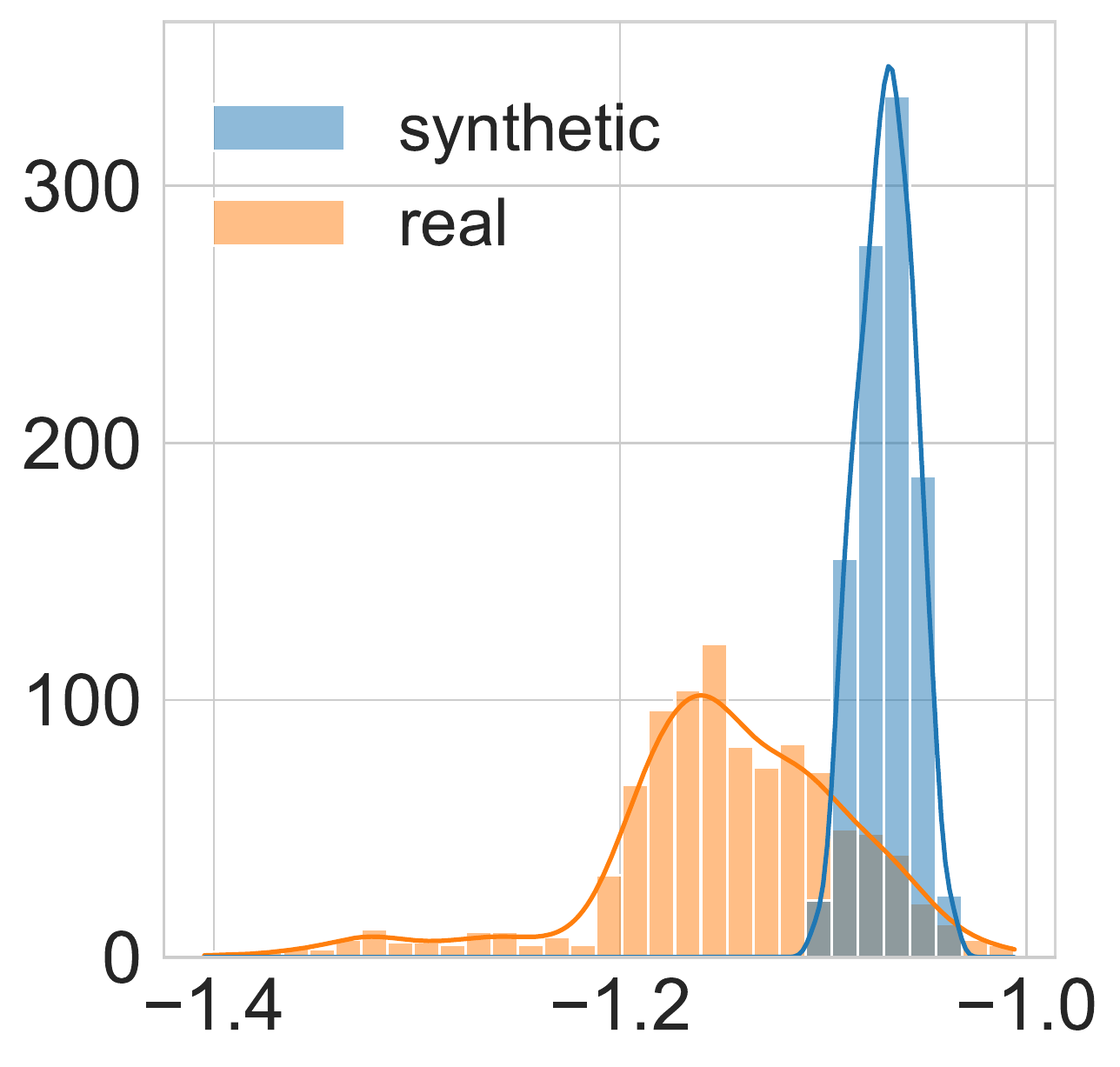}
         \caption{negative\_outlier\_factor}
     \end{subfigure}
     \begin{subfigure}{0.32\textwidth}
         \centering
         \includegraphics[width=\textwidth]{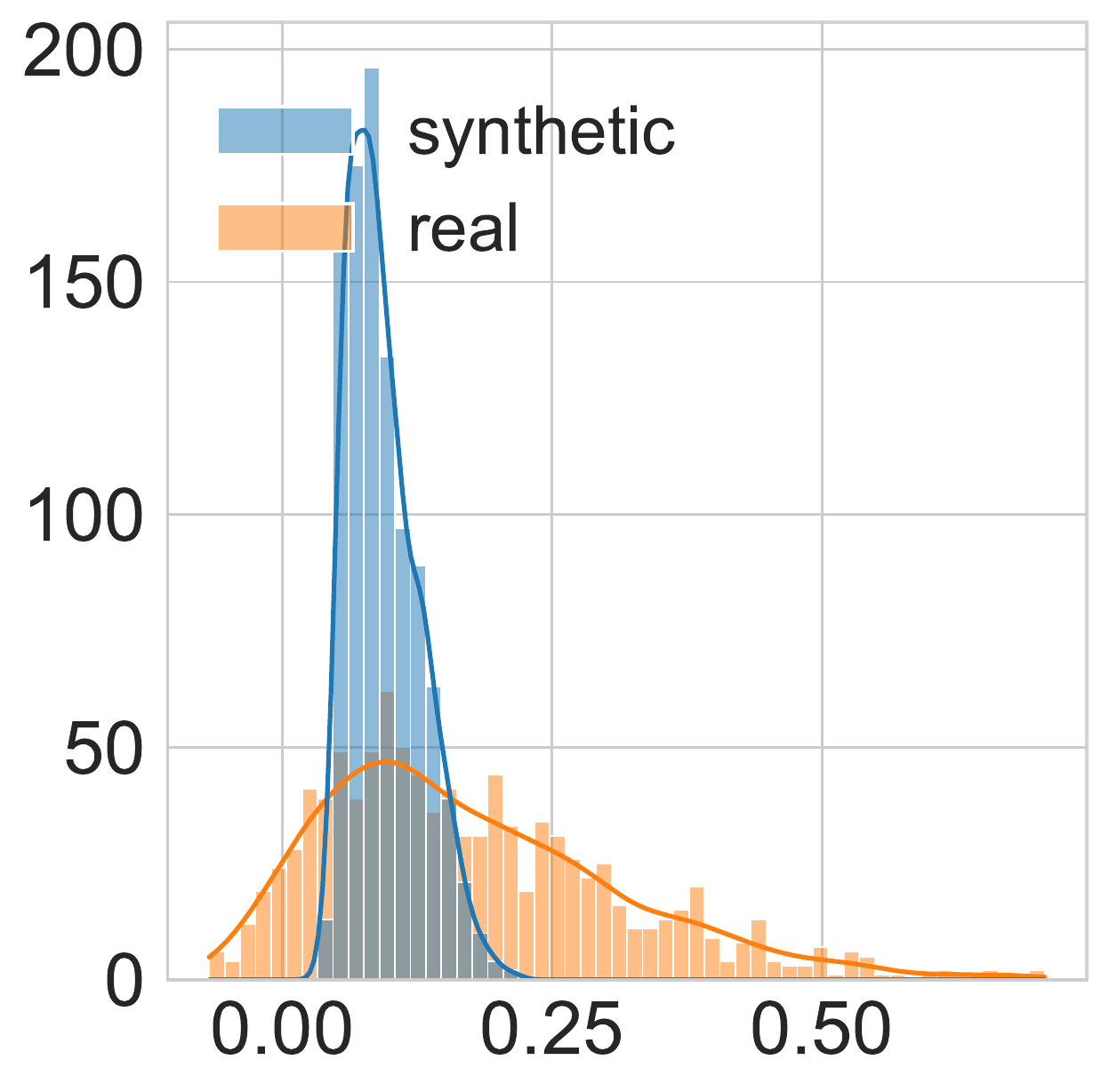}
         \caption{mean\_corr\_ff}
     \end{subfigure}
     \begin{subfigure}{0.32\textwidth}
         \centering
         \includegraphics[width=\textwidth]{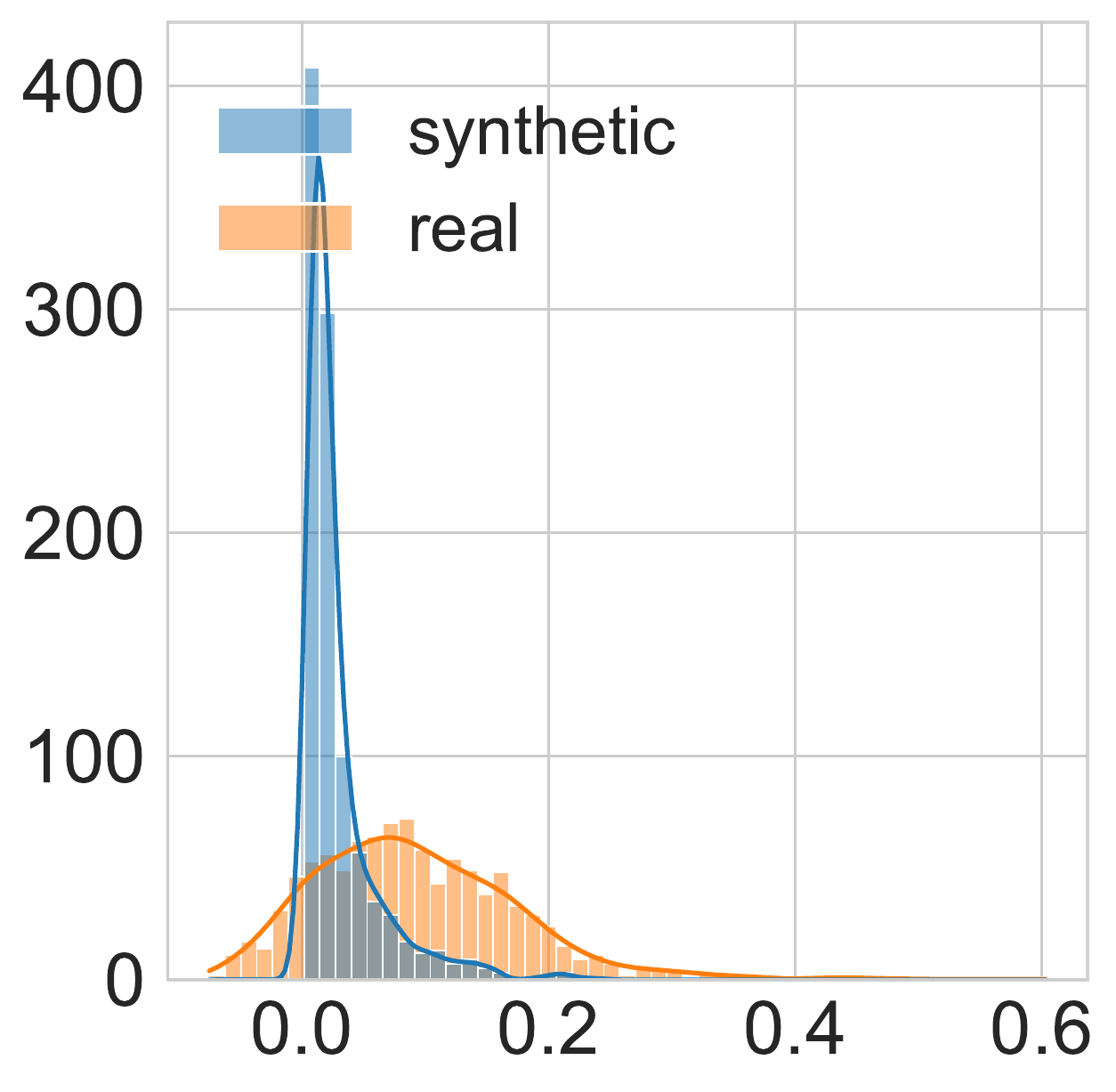}
         \caption{std\_corr\_fc}
     \end{subfigure}

     \begin{subfigure}{0.32\textwidth}
         \centering
         \includegraphics[width=\textwidth]{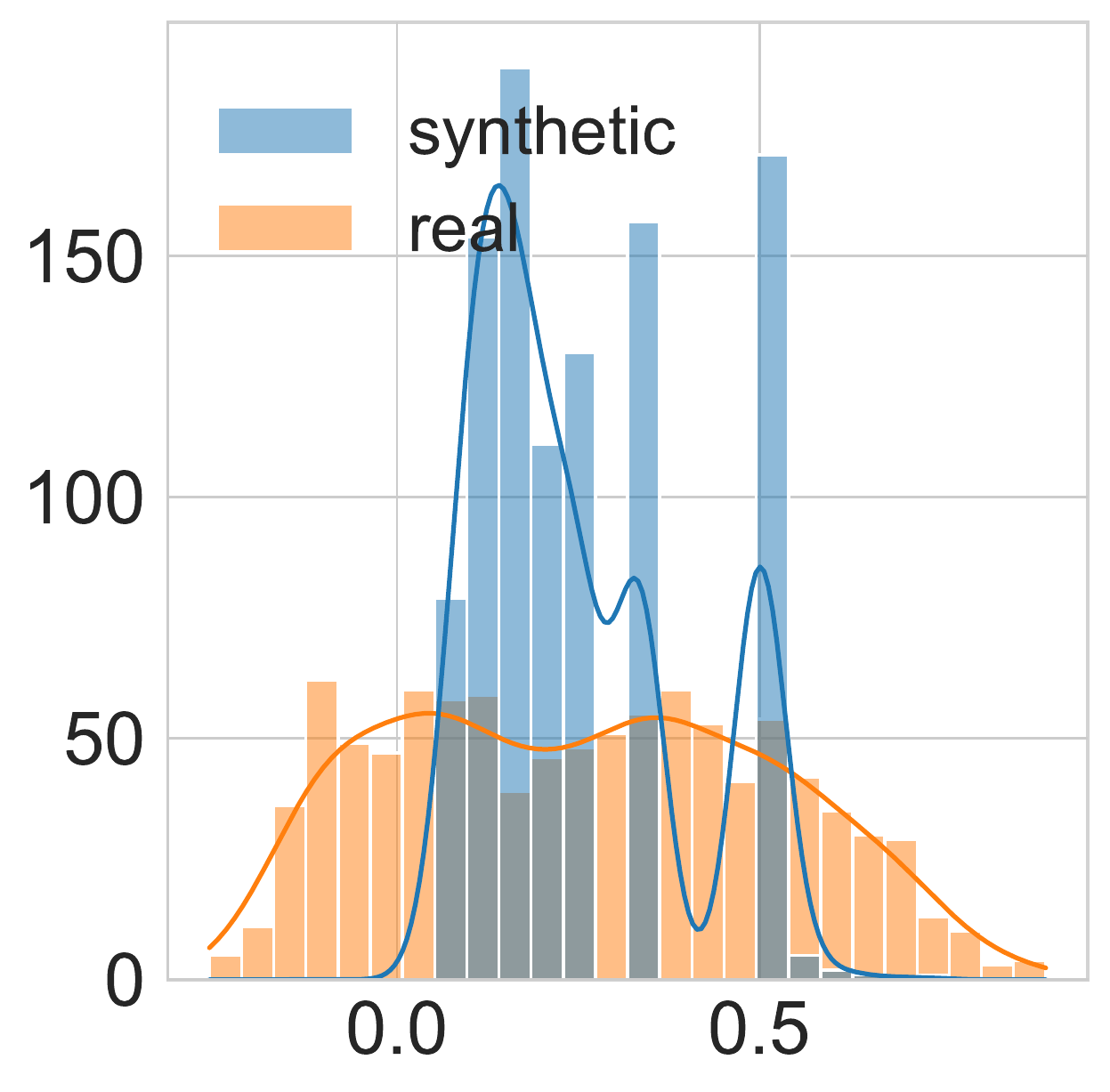}
         \caption{fuzzy\_part\_coeff}
     \end{subfigure}
     \begin{subfigure}{0.32\textwidth}
         \centering
         \includegraphics[width=\textwidth]{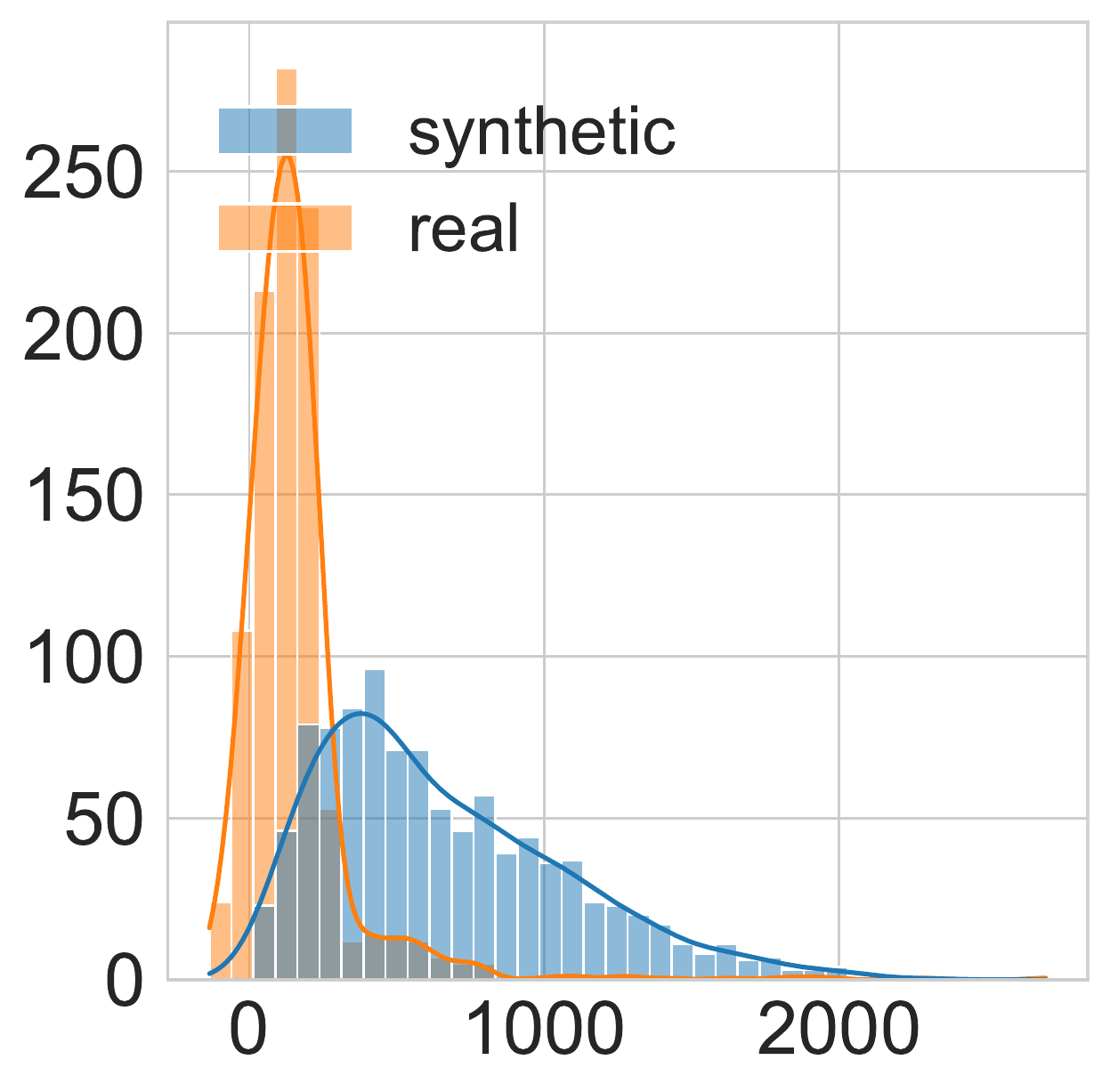}
         \caption{dt\_leaves}
     \end{subfigure}
     \begin{subfigure}{0.32\textwidth}
         \centering
         \includegraphics[width=\textwidth]{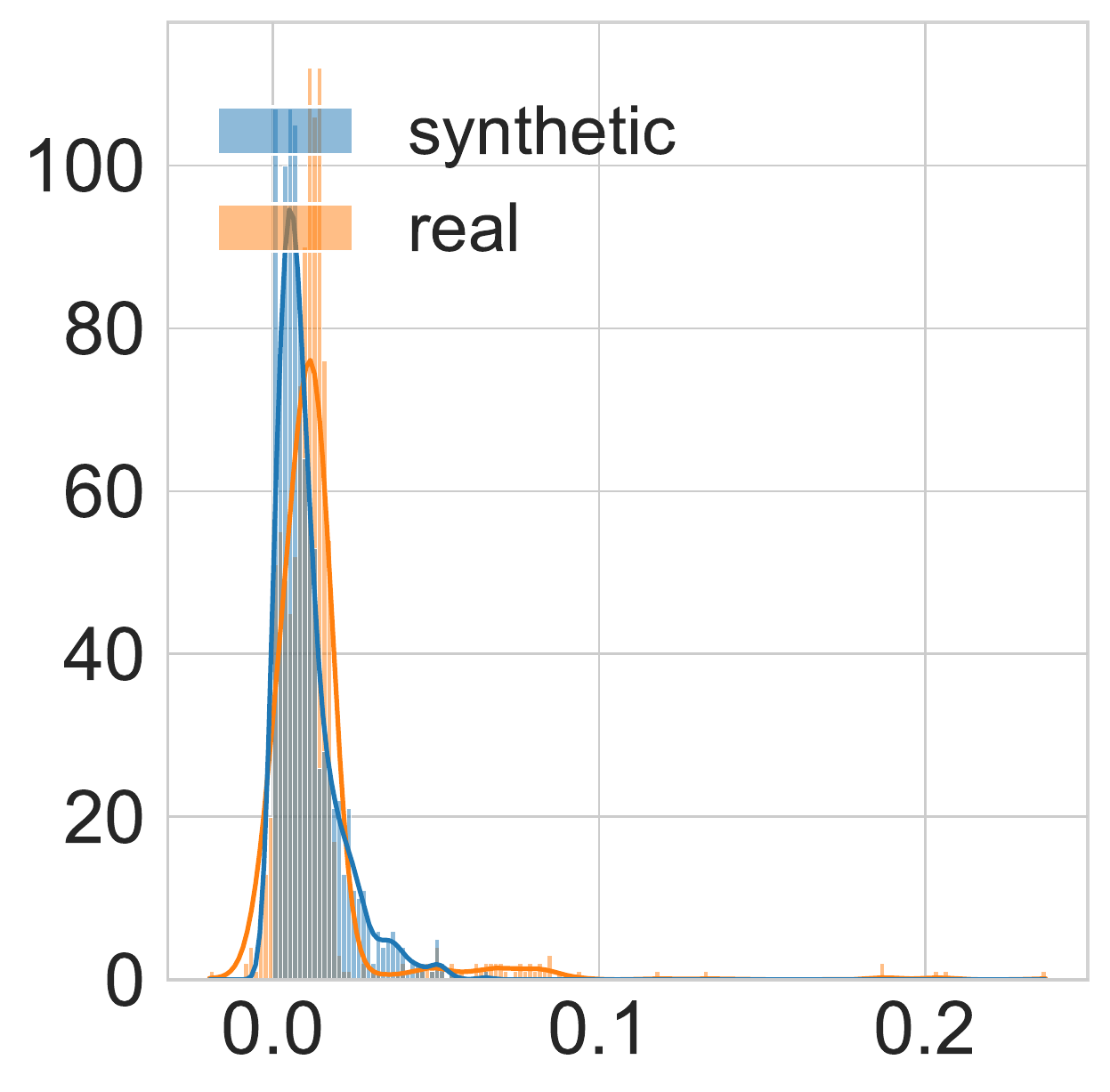}
         \caption{min\_gini\_importance}
     \end{subfigure}
        \caption{Distributions of the most important features according to the synthetic and real datasets for randomly selected samples.}
        \label{fig:distributions}
\end{figure}

It is noticeable that the top-20 meta-features differ significantly when operating with synthetic and real-world data, even when the meta-classifier performs comparably in both cases when it comes to the hit rate. Figure \ref{fig:distributions} shows the distribution of the top-20 meta-features reported in Figure \ref{fig:shap:synthetic} in both the synthetic and real-world data. In this visualization, we have randomly selected 1,000 synthetic instances to facilitate the comparison.

It is clear that the explored meta-features behave differently in both datasets, thus indicating that the synthetic datasets do not resemble real-world datasets. It has not escaped our notice that, in the case of the latter, the top-20 meta-features are related to the data distribution, which is the inner mechanism used by the generative model. On the whole, this suggests that we need stronger approaches to enlarge the training datasets.

\section{Concluding remarks}
\label{sec:conclusions}

This paper presented a meta-classifier to select an optimal model for a structured classification dataset described by its statistical properties. The main novelty of our autoML approach is that it tackles the algorithm selection and hyperparameter tuning problems in a single step. This means that a separate hyperparameter tuning step is not necessary, thus saving processing time and energy. Another contribution concerns a newly introduced set of features that perform better than or equal to the most advanced meta-features reported in the literature. We conjecture that the descriptive power of our compact set of features is given by the approach used to aggregate the multi-valued measures that produce positive and negative values. Instead of computing the minimum, maximum and average across all observations, we process negative and positive values separately, thus avoiding information cancellation. To train the meta-classifier, we generated synthetic data from scratch and from a set of real-world datasets using a generative model.

As for the simulation results, the CNN-based model adapted to deal with structured classification problems emerged as the best-performing meta-classifier. More importantly, our meta-classifier correctly predicted an optimal model for 91\% of the fully synthetic datasets and for 87\% of the datasets generated from real-world problems. As mentioned above, the proposed meta-features played a pivotal role when predicting an optimal model for real-world problems, possibly due to the limited number of training instances available for this setting. In the case of the synthetic datasets, our meta-features and those taken from the literature yield similar results. Finally, we conducted a feature importance analysis to determine which statistical features drive the model selection mechanism. Our findings suggest that the correlation between the problem features, the distribution of decision classes, and the performance of weaker classifiers are suitable proxies for model selection.

While the advantages of our solution were well-supported by the results, there are several improvement points to be addressed in future research steps. Firstly, our proposal will likely suffer from scalability issues as we include more models (classifiers and hyperparameters). This issue can be tackled by a sequential network structure predicting first the most suitable model and next its optimal hyperparameter values. Secondly, the user is forced to define the hyperparameter grid for each model, which might be difficult in the presence of real-valued hyperparameters. This issue can be addressed by modifying the network such that each output neuron represents a hyperparameter, not a hyperparameter value. Consequently, the user would only need to specify the domain of hyperparameters. Thirdly, the differences between the distributions of the synthetic datasets and the real-world datasets make it evident that we need stronger methods for generating training data. 


\bibliographystyle{spmpsci}      
\clearpage

\begin{thebibliography}{10}
\providecommand{\url}[1]{{#1}}
\providecommand{\urlprefix}{URL }
\expandafter\ifx\csname urlstyle\endcsname\relax
  \providecommand{\doi}[1]{DOI~\discretionary{}{}{}#1}\else
  \providecommand{\doi}{DOI~\discretionary{}{}{}\begingroup
  \urlstyle{rm}\Url}\fi

\bibitem{JMLR:meta-features}
Alcobaça, E., Siqueira, F., Rivolli, A., Garcia, L.P.F., Oliva, J.T.,
  de~Carvalho, A.C.P.L.F.: {MFE}: Towards reproducible meta-feature extraction.
\newblock Journal of Machine Learning Research \textbf{21}(111), 1--5 (2020)

\bibitem{ALI2006119}
Ali, S., Smith, K.A.: On learning algorithm selection for classification.
\newblock Applied Soft Computing \textbf{6}(2), 119--138 (2006).
\newblock \doi{10.1016/j.asoc.2004.12.002}

\bibitem{Bello2020}
Bello, M., N\'apoles, G., Sánchez, R., Bello, R., Vanhoof, K.: Deep neural
  network to extract high-level features and labels in multi-label
  classification problems.
\newblock Neurocomputing \textbf{413}, 259--270 (2020).
\newblock \doi{10.1016/j.neucom.2020.06.117}

\bibitem{bensusan2001estimating}
Bensusan, H., Kalousis, A.: Estimating the predictive accuracy of a classifier.
\newblock In: European Conference on Machine Learning, pp. 25--36. Springer
  (2001)

\bibitem{Bezdek1984}
Bezdek, J.C., Ehrlich, R., Full, W.: {FCM}: The fuzzy c-means clustering
  algorithm.
\newblock Computers \& Geosciences \textbf{10}(2), 191--203 (1984).
\newblock \doi{10.1016/0098-3004(84)90020-7}

\bibitem{Brazdil1994}
Brazdil, P., Gama, J., Henery, B.: Characterizing the applicability of
  classification algorithms using meta-level learning.
\newblock In: F.~Bergadano, L.~De~Raedt (eds.) Proceedings of the European
  Conference on Machine Learning (ECML-94), pp. 83--102. Springer Berlin
  Heidelberg, Berlin, Heidelberg (1994)

\bibitem{brazdil2003ranking}
Brazdil, P.B., Soares, C., Da~Costa, J.P.: Ranking learning algorithms: Using
  {IBL} and meta-learning on accuracy and time results.
\newblock Machine Learning \textbf{50}(3), 251--277 (2003)

\bibitem{Cheng2019}
Cheng, Z., Zou, C., Dong, J.: Outlier detection using isolation forest and
  local outlier factor.
\newblock In: Proceedings of the Conference on Research in Adaptive and
  Convergent Systems, RACS '19, p. 161–168. Association for Computing
  Machinery, New York, NY, USA (2019).
\newblock \doi{10.1145/3338840.3355641}

\bibitem{feurer-neurips15a}
Feurer, M., Klein, A., Eggensperger, K., Springenberg, J., Blum, M., Hutter,
  F.: Efficient and robust automated machine learning.
\newblock In: Advances in Neural Information Processing Systems 28 (2015), pp.
  2962--2970 (2015)

\bibitem{godbole2004discriminative}
Godbole, S., Sarawagi, S.: Discriminative methods for multi-labeled
  classification.
\newblock In: Pacific-Asia Conference on Knowledge Discovery and Data Mining,
  pp. 22--30. Springer (2004)

\bibitem{Hospedales2021}
Hospedales, T.M., Antoniou, A., Micaelli, P., Storkey, A.J.: Meta-learning in
  neural networks: A survey.
\newblock IEEE Transactions on Pattern Analysis \& Machine Intelligence
  \textbf{44}(9), 5149--5169 (2021).
\newblock \doi{10.1109/TPAMI.2021.3079209}

\bibitem{Hutter2019}
Hutter, F., Kotthoff, L., Vanschoren, J.: Automated Machine Learning: Methods,
  Systems, Challenges, 1st edn.
\newblock Springer Publishing Company, Incorporated (2019)

\bibitem{jin2019auto}
Jin, H., Song, Q., Hu, X.: {Auto-Keras}: An efficient neural architecture
  search system.
\newblock In: Proceedings of the 25th ACM SIGKDD International Conference on
  Knowledge Discovery \& Data Mining, pp. 1946--1956. ACM (2019)

\bibitem{kalousis1999noemon}
Kalousis, A., Theoharis, T.: {NOEMON}: An intelligent assistant for classi er
  selection.
\newblock In: Proceedings of the ICML-99 Workshop on Recent Advances in
  Meta-Level Learning and Future Work, pp. 28--37 (1999)

\bibitem{Khan2020}
Khan, I., Zhang, X., Rehman, M., Ali, R.: A literature survey and empirical
  study of meta-learning for classifier selection.
\newblock IEEE Access \textbf{8}, 10262--10281 (2020).
\newblock \doi{10.1109/ACCESS.2020.2964726}

\bibitem{SHAP2017}
Lundberg, S.M., Lee, S.I.: A unified approach to interpreting model
  predictions.
\newblock In: I.~Guyon, U.V. Luxburg, S.~Bengio, H.~Wallach, R.~Fergus,
  S.~Vishwanathan, R.~Garnett (eds.) Advances in Neural Information Processing
  Systems 30, pp. 4765--4774. Curran Associates, Inc. (2017).
\newblock \doi{10.5555/3295222.3295230}

\bibitem{Napoles2011}
N\'apoles, G., Bello, M., Salgueiro, Y.: Long-term cognitive network-based
  architecture for multi-label classification.
\newblock Neural Networks \textbf{140}, 39--48 (2021).
\newblock \doi{10.1016/j.neunet.2021.03.001}

\bibitem{padmashani2019rakel}
Padmashani, R., Nivaashini, M., Vidhyapriya, R.: {RAkEL} algorithm and
  mahalanobis distance-based intrusion detection system against network
  intrusions.
\newblock In: International Conference on Artificial Intelligence, Smart Grid
  and Smart City Applications, pp. 689--696. Springer (2019)

\bibitem{Pfahringer2000}
Pfahringer, B., Bensusan, H., Giraud-Carrier, C.G.: Meta-learning by
  landmarking various learning algorithms.
\newblock In: Proceedings of the Seventeenth International Conference on
  Machine Learning, ICML '00, p. 743–750. Morgan Kaufmann Publishers Inc.,
  San Francisco, CA, USA (2000)

\bibitem{pinto2016}
Pinto, F., Soares, C., Mendes-Moreira, J.: Towards automatic generation of
  metafeatures.
\newblock In: Pacific-Asia conference on knowledge discovery and data mining,
  pp. 215--226. Springer (2016)

\bibitem{rakotoarison2022learning}
Rakotoarison, H., Milijaona, L., Rasoanaivo, A., Sebag, M., Schoenauer, M.:
  Learning meta-features for automl.
\newblock In: International Conference on Learning Representations (2022)

\bibitem{reif2014automatic}
Reif, M., Shafait, F., Goldstein, M., Breuel, T., Dengel, A.: Automatic
  classifier selection for non-experts.
\newblock Pattern Analysis and Applications \textbf{17}(1), 83--96 (2014)

\bibitem{Rice1976}
Rice, J.R.: The algorithm selection problem.
\newblock In: Advances in Computers, vol.~15, pp. 65--118. Elsevier (1976)

\bibitem{rivolli2018characterizing}
Rivolli, A., Garcia, L.P., Soares, C., Vanschoren, J., de~Carvalho, A.C.:
  Characterizing classification datasets: a study of meta-features for
  meta-learning.
\newblock arXiv preprint arXiv:1808.10406  (2018)

\bibitem{rivolli2022meta}
Rivolli, A., Garcia, L.P., Soares, C., Vanschoren, J., de~Carvalho, A.C.:
  Meta-features for meta-learning.
\newblock Knowledge-Based Systems \textbf{240}, 108101 (2022)

\bibitem{Ross2014}
Ross, B.C.: {Mutual Information between Discrete and Continuous Data Sets}.
\newblock PLOS ONE \textbf{9}(2), 1--5 (2014).
\newblock \doi{10.1371/journal.pone.0087}

\bibitem{SONG20122672}
Song, Q., Wang, G., Wang, C.: Automatic recommendation of classification
  algorithms based on data set characteristics.
\newblock Pattern Recognition \textbf{45}(7), 2672--2689 (2012).
\newblock \doi{10.1016/j.patcog.2011.12.025}

\bibitem{szymal2019scikit}
Szyma{\'L}, P., Kajdanowicz, T., et~al.: scikit-multilearn: A python library
  for multi-label classification.
\newblock Journal of Machine Learning Research \textbf{20}(6), 1--22 (2019)

\bibitem{Thornton2013}
Thornton, C., Hutter, F., Hoos, H.H., Leyton-Brown, K.: Auto-{WEKA}: Combined
  selection and hyperparameter optimization of classification algorithms.
\newblock In: Proceedings of the 19th ACM SIGKDD Conference on Knowledge
  Discovery and Data Mining (KDD 2013), pp. 847--855 (2013)

\bibitem{Tsoumakas2011}
Tsoumakas, G., Katakis, I., Vlahavas, I.: Random k-labelsets for multilabel
  classification.
\newblock IEEE Transactions on Knowledge and Data Engineering \textbf{23}(7),
  1079--1089 (2011)

\bibitem{vanschoren2018meta}
Vanschoren, J.: Meta-learning: A survey.
\newblock arXiv preprint arXiv:1810.03548  (2018)

\bibitem{Guangtao2014}
Wang, G., Song, Q., Zhang, X., Zhang, K.: A generic multilabel learning-based
  classification algorithm recommendation method.
\newblock ACM Trans. Knowl. Discov. Data \textbf{9}(1) (2014).
\newblock \doi{10.1145/2629474}

\bibitem{wang2015improved}
Wang, G., Song, Q., Zhu, X.: An improved data characterization method and its
  application in classification algorithm recommendation.
\newblock Applied Intelligence \textbf{43}(4), 892--912 (2015)

\bibitem{xu2019modeling}
Xu, L., Skoularidou, M., Cuesta-Infante, A., Veeramachaneni, K.: Modeling
  tabular data using conditional {GAN}.
\newblock Advances in Neural Information Processing Systems \textbf{32} (2019)

\bibitem{zhang2018binary}
Zhang, M.L., Li, Y.K., Liu, X.Y., Geng, X.: Binary relevance for multi-label
  learning: an overview.
\newblock Frontiers of Computer Science \textbf{12}(2), 191--202 (2018)

\bibitem{zhang2007ml}
Zhang, M.L., Zhou, Z.H.: {ML-KNN}: A lazy learning approach to multi-label
  learning.
\newblock Pattern recognition \textbf{40}(7), 2038--2048 (2007)

\bibitem{zhu2020}
Zhu, D., Zhu, H., Liu, X., Li, H., Wang, F., Li, H., Feng, D.: {CREDO}:
  Efficient and privacy-preserving multi-level medical pre-diagnosis based on
  ml-knn.
\newblock Information Sciences \textbf{514}, 244--262 (2020)

\bibitem{ZHU2018171}
Zhu, X., Yang, X., Ying, C., Wang, G.: A new classification algorithm
  recommendation method based on link prediction.
\newblock Knowledge-Based Systems \textbf{159}, 171--185 (2018).
\newblock \doi{10.1016/j.knosys.2018.07.015}

\bibitem{Zimmer2021}
Zimmer, L., Lindauer, M., Hutter, F.: {Auto-PyTorch} tabular: Multi-fidelity
  metalearning for efficient and robust {AutoDL}.
\newblock IEEE Transactions on Pattern Analysis and Machine Intelligence pp.
  3079--3090 (2021)

\end{thebibliography}

\end{document}